\documentclass[acmlarge]{acmart}

\AtBeginDocument{%
  }

\usepackage{natbib}  
\usepackage{caption} 

\usepackage{algorithm}
\usepackage{algorithmic}
\usepackage{xcolor}
\usepackage{subcaption}
\usepackage{wrapfig}
\usepackage{multirow}
\usepackage{makecell}
\usepackage{amsmath}
\usepackage{enumitem}

\newcommand{\cA}{\mathcal{A}}  
\newcommand{\cC}{\mathcal{C}} \newcommand{\cD}{\mathcal{D}}

\newcommand{\cM}{\mathcal{M}}

\newcommand{\cU}{\mathcal{U}}

\newcommand{\MG}{\cM^G_\theta}
\newcommand{\MP}{\cM^{P^a_i}_\theta}

\newcommand{\MC}{\cM^{C^a_i}_\theta}
\newcommand{\Ui}{\cU_i}
\newcommand{\Ca}{\cC_a}
\newcommand{\Cu}{\cC_u}
\newcommand{\Dai}{\cD^a_i}
\newcommand{\Dui}{\cD^u_i}

\usepackage{pifont}
\newcommand{\cmark}{\ding{51}}%
\newcommand{\xmark}{\ding{55}}%

\DeclareMathOperator*{\argmin}{\textit{argmin }}

\begin{document}

\title{CRoP: Context-wise Robust Static Human-Sensing Personalization}

\author{Sawinder Kaur}
\email{sakaur@syr.edu}
\orcid{0009-0002-8937-3702}
\affiliation{%
  \institution{Syracuse University}
  \city{Syracuse}
  \state{New York}
  \country{USA}
}

\author{Avery Gump}
\email{averygmp@gmail.com}
\orcid{0009-0007-9535-3974}
\affiliation{%
  \institution{University of Wisconsin-Madison}
  \city{Madison}
  \state{Wisconsin}
  \country{USA}
}

\author{Yi Xiao}
\affiliation{%
  \institution{Arizona State University}
  \city{Tempe}
  \state{Arizona}
  \country{USA}}
\email{yxiao124@asu.edu}
\orcid{0000-0002-5261-5440}

\author{Jingyu Xin}
\affiliation{%
  \institution{Syracuse University}
  \city{Syracuse}
  \state{New York}
  \country{USA}
}
\email{jxin05@syr.edu}
\orcid{0000-0001-6309-6845}

\author{Harshit Sharma}
\affiliation{%
  \institution{Arizona State University}
  \city{Tempe}
  \state{Arizona}
  \country{USA}}
\email{hsharm62@asu.edu}
\orcid{0000-0002-7016-6220}

\author{Nina R Benway}
\email{nrbenway@syr.edu}
\orcid{0000-0003-0955-9495}
\affiliation{%
  \institution{University of Maryland-College Park}
  \city{College Park}
  \state{Maryland}
  \country{USA}}

\author{Jonathan L Preston}
\email{jopresto@syr.edu}
\orcid{0000-0001-9971-6321}
\affiliation{%
  \institution{Syracuse University}
  \city{Syracuse}
  \state{New York}
  \country{USA}
}

\author{Asif Salekin}
\email{asif.salekin@asu.edu}
\orcid{0000-0002-0807-8967}
\authornote{Corresponding author}
\affiliation{%
  \institution{Arizona State University}
  \city{Tempe}
  \state{Arizona}
  \country{USA}}

\renewcommand{\shortauthors}{Kaur et al.}

\begin{abstract}
  The advancement in deep learning and internet-of-things have led to diverse human sensing applications. However, distinct patterns in human sensing, influenced by various factors or contexts, challenge the generic neural network model's performance due to natural distribution shifts. To address this, personalization tailors models to individual users. Yet most personalization studies overlook intra-user heterogeneity across contexts in sensory data, limiting intra-user generalizability.  This limitation is especially critical in clinical applications, where limited data availability hampers both generalizability and personalization. Notably, intra-user sensing attributes are expected to change due to external factors such as treatment progression, further complicating the challenges.
  To address the intra-user generalization challenge, this work introduces \textcolor{violet}{CRoP}, a novel static personalization approach. \textcolor{violet}{CRoP} leverages off-the-shelf pre-trained models as generic starting points and captures user-specific traits through adaptive pruning on a minimal sub-network while allowing generic knowledge to be incorporated in remaining parameters.  \textcolor{violet}{CRoP} demonstrates superior personalization effectiveness and intra-user robustness across four human-sensing datasets, including two from real-world health domains, underscoring its practical and social impact.
  Additionally, to support \textcolor{violet}{CRoP}'s generalization ability and design choices, we provide empirical justification through gradient inner product analysis, ablation studies, and comparisons against state-of-the-art baselines.
\end{abstract}

\ccsdesc[500]{Human-centered computing~Ubiquitous and mobile computing theory, concepts and paradigms}

\keywords{Intra-user Generalization, Personalization, Context-wise Robustness, Robust Personalization, Adaptive Personalization}

\maketitle
\section{Introduction}
\label{introduction}
Recent automated human sensing applications—like activity recognition, fall detection, and health tracking - revolutionize daily life, especially in personal health management \citep{Wang2023-zi}. However, unique user patterns and natural distribution shifts \citep{gong2023dapper} caused by behaviors, physical traits, environment, and device placements \citep{ustev2013user, stisen2015smart} lead to the underperformance of generic sensing models in practical use. To tackle this, various domain adaptation techniques have been explored, with personalization widely used to adapt a generic model to the target user's specific domain \citep{ meegahapola2023generalization, sempionatto2021wearable}. In literature, personalization occurs either during the enrollment phase (static) \citep{EMGSense,MobilePhys,PTN-baseline}  or continuously throughout application use, a process known as continual learning \citep{daniels2023efficient,wu2024testtime}. 


Continual learning methods involve retraining models with new data, either supervised or unsupervised. While these approaches enable models to adapt to new patterns and changes in data distribution over time, they often face efficiency challenges. As \citet{inefficient-cl} points out, these methods, particularly those that prevent catastrophic forgetting \cite{daniels2023efficient,cassle,kaizen,Tang2024BalancingCL,packnet,Piggyback,cotta}, often struggle with memory, computation, and storage requirements inefficiencies, limiting their real-world applicability. Frequent model updates on user devices can introduce significant delays. This is especially true for devices with limited processing power, such as smartphones or wearables. These delays reduce the responsiveness of the application. They also lead to battery drain and inefficiency in real-time applications. Furthermore, to avoid catastrophic forgetting, several continual learning approaches rely on replay-based methods which require storage of previously encountered data \cite{vandeven2019generativereplayfeedbackconnections,hayes2019memoryefficientexperiencereplay,Van-de-Ven2020-jh}. This not only increases storage requirements but also raises privacy concerns. Constantly storing potentially identifiable information increases the risk of data breaches or misuse. This is especially concerning in regulatory environments, such as healthcare \cite{privacy-cl}.

Additionally, supervised continual learning approaches face the significant challenge of requiring expert-labeled data for each new batch of user interactions, also termed as label delay \cite{csaba2024labeldelayonlinecontinual}. This demand for continuous, high-quality labels can make it impractical for many applications, especially in settings where expert annotation is costly or unavailable. For example, in clinical or health monitoring contexts, each new data point may need to be validated by professionals, which is time-consuming and difficult to scale.


In contrast, static personalization offers an efficient alternative by customizing the model with a one-time, limited dataset collected during enrollment. This approach minimizes computation, requires no ongoing data or label storage or collection, and reduces user engagement, making it especially suitable for resource-constrained human-sensing applications.
However, existing such studies often overlook intra-user variability due to factors like changes in magnetic field \citep{effect-mf}, sensor position \citep{PARK201431}, terrain \citep{Kowalsky2021-ox}, or the health symptoms \cite{Paeske2023-eb}, leading to poor intra-user generalizability for contexts not present during personalization. For instance, a smartphone activity recognition model personalized with handheld data may perform poorly when the phone is in a pocket. Notably, the distribution of clinical data is expected to shift, even within the same individual. For instance, in clinical speech technologies, changes in data distribution over time may occur due to the progression of neurodegenerative diseases, relevant for disease monitoring apps \cite{Stegmann2020-ot}, or through desired learning mechanisms resulting from the use of technology, as seen in automated speech therapy apps \cite{benway-and-preston}.


This research gap is worsened since static personalization typically relies on a small sample set from the target user, covering limited available contexts—particularly in clinical settings or applications with data scarcity \cite{Berisha2021-mx,benway-and-preston}. Commercial human sensing technologies like Google Assistant, Amazon Alexa, and Apple's Siri also statically personalize speech recognition models using limited phrases \citep{hey-siri-personalization, google-assistant-personalization, amazon-speech-training}. Similarly, the Apple Watch uses initial calibration for enhanced running activity tracking \citep{apple--support,apple-activity-training}. This limited available context during personalization is problematic, as shown in Section \ref{motivation}, where we demonstrate that static personalization may improve performance in the available training contexts but can also significantly degrade it in other unseen contexts for the same user.


Therefore, given the importance of static personalization in human sensing, this paper addresses its intra-user generalizability gap. Since personalized models are tailored to individual users and not intended for use by others, inter-user generalizability is outside the scope of this work.
An additional challenge is that several personalization approaches in the literature, such as EMGSense \cite{EMGSense} and MobilePhys \cite{MobilePhys}, train their own generic models, sometimes even using data from target users. This limits privacy-preserving generic model sharing, and, in some cases, requires target users to share sensitive data, raising additional privacy concerns. 
It also complicates adding a new user, as the generic model must be updated and redistributed. While model redistribution is common in federated learning, human-sensing personalization isn’t limited to federated learning methods. To address this limitation, this paper only uses pre-trained, off-the-shelf models as generic models, which do not require data from the target users.


\begin{wrapfigure}{r}{0.5\textwidth}
\vskip -3ex
\includegraphics[width=0.99\linewidth]{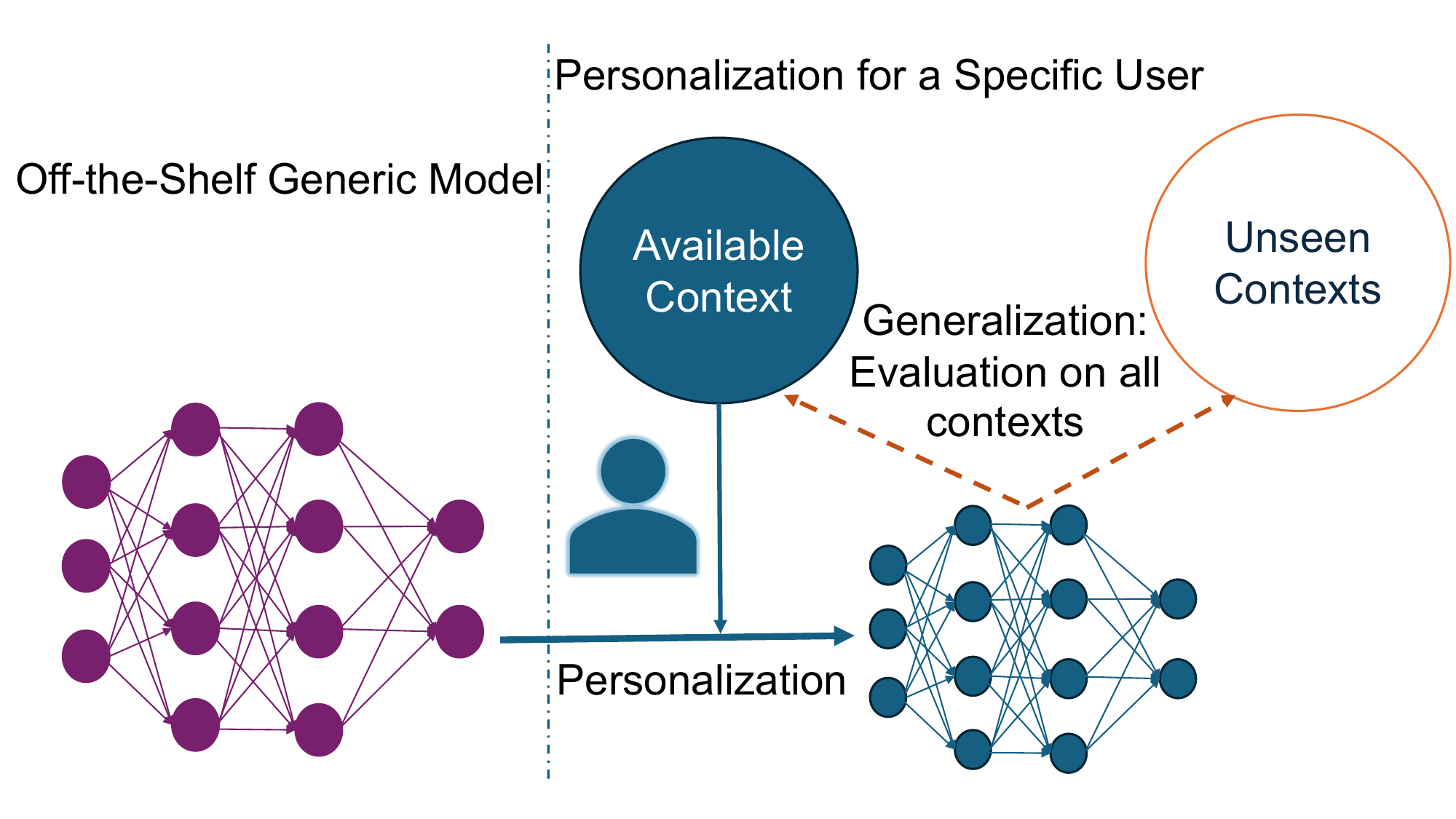}
        \caption{Problem Setting: Study Objective}
    \label{fig:main-figure}
\vskip -3ex
\end{wrapfigure}

In summary, Figure \ref{fig:main-figure} outlines the paper's scope and objective. The goal is to develop an intra-user robust approach to personalize an off-the-shelf generic model for a specific user using limited data from limited available contexts. The primary objective is to ensure that the personalized model thus obtained exhibits robust generalization capabilities across all contexts (i.e., available and unseen contexts). Crucially, unseen context data takes no part in training or adjusting the personalized model outcome, and the generic model remains entirely off-the-shelf, with no accessibility for modification or design choices. These constraints highlight the real-world impact of this research, particularly in clinical settings where privacy concerns often limit data sharing \cite{malin2018between,Rathbone2023-sy}, and only trained off-the-shelf models are shared among researchers and developers.

To achieve the research objective in Figure \ref{fig:main-figure}, this paper introduces \textcolor{violet}{CRoP}, a novel approach to create context-wise intra-user robust static personalized models. The key contributions are:

\begin{enumerate}[topsep=0pt,nolistsep, leftmargin=*]
    \item It facilitates utilizing readily available off-the-shelf pre-trained models with state-of-the-art accuracy, eliminating the need for training customized generic models, thus reducing training effort and providing a strong foundation for personalization.

    \item 
    
     

     Personalization with intra-user generalization has two challenges: i) learning user-specific patterns and ii) keeping information about generic patterns intact. \textcolor{violet}{CRoP} addresses these challenges by leveraging adaptive pruning and regularization followed by a mixing step to optimize this balance. After initial finetuning or model adaptation for the target user, \textcolor{violet}{CRoP} (its \emph{ToleratedPrune} module) dynamically adapts pruning levels per user to extract a personalized substructure that best represents the user’s traits in the available context. When users require significant parameter adjustments, \textcolor{violet}{CRoP}’s regularizer increases parameter polarization, enabling more aggressive pruning and restoration of more parameter values, i.e., information from the generic model in the subsequent mixing step. This adaptive pruning strategy dynamically balances personalization and generalization by tailoring the recovery of generic information to user needs.
    \item To showcase \textcolor{violet}{CRoP}'s efficacy, comprehensive evaluations were performed on four human sensing datasets: PERCERT-R \cite{Benway2023-eu}: a clinical speech therapy dataset, WIDAR \citep{WIDAR-dataset}: a lab-based WiFi-CSI dataset, ExtraSensory: a real-world mobile sensing dataset \citep{Extrasensory-dataset}, and a stress-sensing dataset \citep{stress-sensing}, while considering two disjoint contexts for each dataset. In order to obtain information about context variation within PERCEPT-R and Stress-sensing, we collaborated with the original authors of these two datasets. This additional annotation will be released alongside this publication.
    \item On average across all datasets and corresponding all contexts, \textcolor{violet}{CRoP}  achieves a 35.23 percentage point improvement in personalization compared to the generic model, and a 7.78 percentage point improvement in generalization compared to the \emph{conventional-finetuned} model \citep{HAR-P}. As compared to the best baseline (Packnet), these gains are 9.18 and 9.17 percent points higher, respectively. Moreover, alongside a detailed ablation study discussion in Section \ref{ablation-study}, an empirical justification of \textcolor{violet}{CRoP}’s design choices that enable intra-user generalizability among different contexts is provided in Section \ref{sec:empirical-justification}, employing Gradient Inner Product(GIP) \citep{shi2021gradient} analysis. Additionally, in order to demonstrate the feasibility of on-device personalization through \textcolor{violet}{CRoP}, we compute training time and resource requirements for seven platforms or devices. 
\end{enumerate}

\section{Related Work}
\label{related-work}

This section summarizes and critiques key state-of-the-art approaches with similar objectives to those addressed in this work, highlighting their contributions and limitations in practical, user-centered adaptation settings.

\subsection{Domain Adaptation and Generalization}
\label{related-work-da}
Domain adaptation (DA) and Domain Generalization (DG) techniques address performance drops due to distributional shifts between source and target domains \cite{kouw2019introductiondomainadaptationtransfer}. These methods are useful when models trained on generic data need to adapt to new users or contexts. The key difference is that while DA approaches have access to target domain data, DG techniques work without any target domain data, not even unlabeled \cite{dg-survey}. Thus, the scenario addressed by DA approaches is more similar to our problem setting.
Domain adaptation approaches typically rely on target domain data \cite{7452659} or require access to source data without specialized training of the generic model \cite{9080115,10.5555/3045118.3045244}. Source-free domain adaptation (SFDA) methods, such as SHOT \citep{liang2020we}, eliminate the need for source data or retraining the generic model. SHOT transfers the source hypothesis by freezing classifier weights and fine-tuning the feature extractor using self-supervised pseudo-labeling to align target representations.

Given that the problem addressed by SHOT aligns with the criteria of having no access to generic data and avoiding specialized generic model training, we adopted SHOT as one of the baseline approaches in this study. However, it is important to note that SHOT does not effectively tackle the limitations posed by restricted-context data during the fine-tuning process. SHOT employs pseudo-labels generated using a combination of k-means clustering and nearest neighbor classification on target domain's data. In our setting, that target  domain would be user's available context data. Since, SHOT's training approach does not consider futher changes in the target domain, SHOT would require another round of training using unseen domain's data which is unavailable in current setting. As a result, its performance may not adequately generalize to the intra-user variability present in the data, which is a crucial aspect of our research focus.

\subsection{Continual Learning}

Continual personalization approaches \cite{daniels2023efficient,cassle,kaizen,Tang2024BalancingCL,packnet,Piggyback,cotta} can improve intra-user generalizability by continually fine-tuning the model as new data arrives.
Some of these approaches \cite{cassle,kaizen,Tang2024BalancingCL} require specialized training of the generic model, limiting the use of off-the-shelf pre-trained models, while others \citep{packnet,Piggyback} require continued stream of labeled data, limiting their application in health-care scenarios.  However, all continuous learning approaches require repeated computation overhead to adjust the model outcome to new data \citep{10204648}, which can be infeasible in real-time applications, more so for resource-constrained platforms like wearables \cite{10.1145/3242969.3242985}, which is prominent for health sensing such as stress or fall detection. 



Packnet \cite{packnet} and Piggyback \cite{Piggyback} adapt the generic model for a stream of continuously changing tasks. Packnet relies on finetuning, pruning and re-training the model for each new task. On the other hand, Piggyback employs pruning to learn a new binary mask for each new task which is then applied to a generic model in order to get task-specific results. However, this requires correct identification of the task before choosing the mask. 


Notably, all continual learning approaches, including PackNet and Piggyback, require target domain data to adapt, limiting their ability to handle entirely unseen domains. While they mitigate catastrophic forgetting, they focus on previously encountered contexts. In contrast, our framework excludes unseen context data during adaptation, exposing a key limitation of existing methods.


\subsection{Test-Time Adaptation (TTA)}
\label{related-work-tta}


Test-time adaptation offers a more efficient and flexible alternative, allowing models to adapt in real-time during deployment 
by making adjustments based on incoming data without necessitating a complete retraining process. By focusing on adapting the model in real-time, test-time adaptation minimizes resource consumption as compared to continual learning approaches. While \citet{wu2024testtime} achieves test-time domain adaptation by manipulating batch-normalization layers, thus enforcing a restriction on model architecture. However, Continual Test Time Domain Adaptation (CoTTA) \cite{cotta} does not have such restrictions and allows the use of off-the-shelf models.


CoTTA \cite{cotta} is an unsupervised learning approach designed to enhance model adaptation in dynamic environments. It utilizes a teacher model, which is initially set up as a replica of the generic model, to generate pseudo-labels for training. During each iteration, the teacher model is updated using weighted summation of current states of the teacher and the student model. This allows the teacher model to generate pseudo-lables which align with the evolving data patterns over time. Additionally, the current model state is subject to a stochastic restoration of weights, implemented through a Bernoulli distribution with a very low probability of success. This randomization introduces an element of variability that helps prevent overfitting and encourages exploration of the weight space, further enhancing the model's ability to generalize to unseen data. 


\begin{wrapfigure}{r}{0.2\textwidth}
    \vspace{-2ex}
    \centering
    \includegraphics[width=0.9\linewidth]{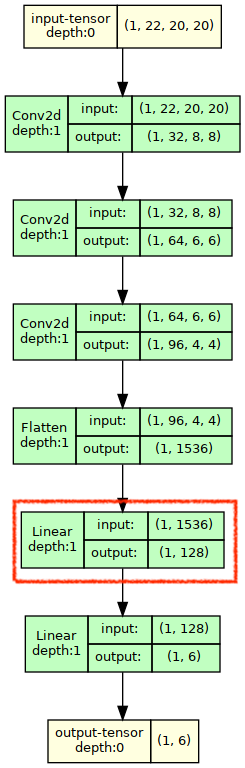}
    \vspace{-2ex}
    \caption{Model Structure for the Lenet Model used for Widar Dataset. The parameters of the highlighted layer are used for preliminary analysis}
    \label{fig:lenet-structure}
\vspace{-8ex}
\end{wrapfigure}

We adopt CoTTA as a baseline due to its similarity to our problem setting, where available context data supports test-time adaptation, and unseen context data is reserved for evaluation. However, test-time adaptation methods like CoTTA increase inference resource demands, potentially causing delays that hinder real-time usability—a key limitation in time-sensitive applications.

\subsection{Personalization Approaches}


 A few static personalization approaches \citep{PTN-baseline,EMGSense,MobilePhys} aimed for the additional goal of out-of-distribution robustness. However, these methods require access to the generic model—either to make specific design choices \citep{PTN-baseline}, which prevents them from utilizing off-the-shelf models, or to incorporate knowledge about the target user's data distribution during the generic model's training phase \citep{EMGSense,MobilePhys}, raising privacy concerns.

Recently, \citet{PTN-baseline} extends the idea of triplet loss \citep{FaceNet} to personalization (PTN). The optimization objective combines the minimization of the Euclidean distance between data from the same target classes while maximizing the Euclidean distance between different target classes in order to learn an embedding optimized for the desired task. The features extracted using DNN have been shown to be superior to the engineered features \citep{10.1007/978-3-319-63558-3_40}. Thus, \citet{PTN-baseline} uses these features to train a KNN for the final prediction task. The approach has been shown to be robust for out-of-distribution data.

However, none of these approaches address intra-user variability in data. Since PTN allows the usage of off-the-shelf model and does not require sharing of the personal data for generic model training, we consider PTN as one of the baselines.

\subsection{Selecting Baselines}
The problem addressed in this work has the following requirements: using an Off-the-shelf generic model, having no access to the generic data, and requiring adaptation to available context data and robustness to unseen data. As discussed in the previous sections, the approaches SHOT \cite{liang2020we}, PTN \cite{PTN-baseline}, PackNet \cite{packnet}, Piggyback \cite{Piggyback}, and Continual Test Time Domain Adaptation (CoTTA) \cite{cotta} comply with the first three requirements. Thus, we adapt these approaches to the problem statement addressed in this work and use them as our baselines. These approaches typically rely on training data from a specific context to adapt to that context. However, the scenario addressed in this work limits training data to only one context, with the expectation that the approach will also perform well in unseen contexts. Since data from unseen contexts is unavailable, we prevent these approaches from using it during training.
Instead, the data used for training or adaptation is restricted to data belonging to the available context only, while the unseen context data is used merely for testing.

\begin{wrapfigure}{r}{0.35\textwidth}
    \centering
    \vspace{-10ex}
    \begin{subfigure}[b]{\linewidth}
\includegraphics[width=0.90\linewidth]{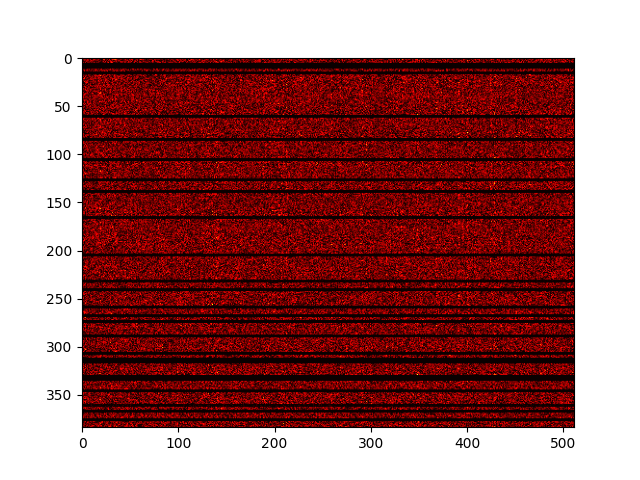}
         \caption{Context C1}
    \end{subfigure} 
    \begin{subfigure}[b]{\linewidth}\includegraphics[width=0.90\linewidth]{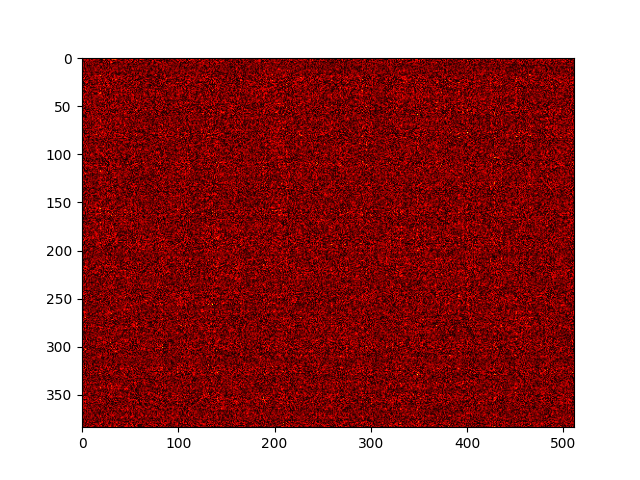}
         \caption{Context C2}
    \end{subfigure}
    \caption{Heat map for the absolute magnitude of parameters belonging to penultimate layer for LeNet models finetuned using data from context (a) C1 and (b) C2. The first linear layer after the convolution layer has been used for this analysis. The size of the original weight matrix of the layer is 1536 X 128. However, we reshape heat map to 384 X 512 for better observability.}
    \vskip -2ex
    \label{fig:heat-map}
\end{wrapfigure}

\section{Motivation}
\label{motivation}

When learning patterns from human sensing data in a limited context, conventional fine-tuning approaches can overwrite generic knowledge that is not relevant to that specific context but applicable to others, leading to a performance drop in those unrepresented contexts. To illustrate this, we conducted a preliminary study comparing the performance of generic and \emph{conventionally-finetuned} \citep{HAR-P} personalized human-gesture-recognition models using the LeNet architecture \citep{WIDAR-dataset} trained on the WIDAR dataset. Data preprocessing details are discussed in the Section \ref{sec:experiments} (Experiments).

\begin{table}[]
    \centering
    \subfloat[Context C1]{
\resizebox{0.4\linewidth}{!}{
    \begin{tabular}{|c|cc|cc|cc|}
        \hline
        Model& \multicolumn{2}{|c|}{Generic} & \multicolumn{2}{|c|}{Personalized} &\multicolumn{2}{|c|}{$\Delta$}\\
        \hline
         User& C1 & C2 &  C1 & C2 & C1 & C2 \\
         \hline
         0&63.90&77.09&87.06&65.02&\textcolor{cyan}{+23.16}&\textcolor{orange}{-11.88}\\
         1&61.80&79.78&89.38&44.38&\textcolor{cyan}{+27.57}&\textcolor{orange}{-35.40}\\
         2&45.63&79.81&71.88&64.45&\textcolor{cyan}{+29.75}&\textcolor{orange}{-26.62}\\
         \hline
         Average&&&&&\textcolor{cyan}{+26.82}& \textcolor{orange}{-24.63}\\
         \hline
    \end{tabular}}
    \label{tab:Global-vs-personalized}}
\quad
\subfloat[Context C2]{
\resizebox{0.4\linewidth}{!}{
    \begin{tabular}{|c|cc|cc|cc|}
        \hline
        Model& \multicolumn{2}{|c|}{Generic} & \multicolumn{2}{|c|}{Personalized} &\multicolumn{2}{|c|}{$\Delta$}\\
        \hline
         User& C1 & C2 &  C1 & C2 & C1 & C2 \\
         \hline
         0&60.80&73.28&57.46&77.30&\textcolor{orange}{-3.34}&\textcolor{cyan}{+4.02}\\
         1&59.58&73.18&42.38&93.75&\textcolor{orange}{-17.29}&\textcolor{cyan}{+20.57}\\
         2&46.13&80.46&40.19&87.15&\textcolor{orange}{-5.93}&\textcolor{cyan}{+6.69}\\
         \hline
         Average&&&&&\textcolor{orange}{-8.85}& \textcolor{cyan}{+10.43}\\
         \hline
    \end{tabular}}
    \label{tab:Global-vs-personalized-2}
    }
    \caption{Inference accuracy comparison of generic model with conventionally trained personalized model}
    \vspace{-8ex}
\end{table}


\par 
Table \ref{tab:Global-vs-personalized} compares the performance of generic and \emph{conventionally-finetuned} \citep{HAR-P} personalized models on each user’s data belonging to context C1 and evaluated to the same user’s disjoint data in both C1 (available) and C2 (unseen) contexts. It can be observed that conventional finetuning introduces a significant gain of $26.82\%$ for context C1's data but at the cost of $24.63\%$ reduction in context C2. Similar patterns are seen when personalization is performed on context C2 as shown in Table \ref{tab:Global-vs-personalized-2}. 
Thus, conventional finetuning-based personalization using limited data from one context can significantly worsen the model's performance in an unseen context.

To investigate this discrepancy in performance, we compare the distribution of parameter magnitudes of the models personalized on contexts C1 and C2 using \emph{conventional finetuning}, as shown using heat maps in Figures \ref{fig:heat-map} (a) and \ref{fig:heat-map} (b). Notably, both models started from the same generic model but were personalized independently, leading to different parameter magnitudes tailored to the traits of two different contexts. The structure of the Lenet Model used for the Widar Dataset is provided in Figure \ref{fig:lenet-structure}.  The first linear layer after the convolution layers, that is, `fc1', is chosen for this analysis as it captures the relative importance of input features \citep{Goodfellow2016}. Although the original size of the layer is 1536 X 128, we resized the corresponding heatmap to 384 X 512 to improve the visualization. In the heatmaps, the brighter shades of red represent a higher magnitude, while the black pixel represents a weight magnitude almost equal to zero. Thus, if a feature is considered less important, all corresponding parameters (connections) will have lower magnitudes, appearing as black rows in the heat map. Notably, there is a substantial difference in parameter magnitudes (i.e., color differences) between models trained in different contexts. Specifically, many rows in Figure \ref{fig:heat-map}(a) have black pixels, representing less useful features. These pixels represent parameters with magnitudes close to zero (empirically identified as lower than $1X10^{-4}$), and they comprise $18.66\%$ of the first linear layer for the model personalized on context C1. However, the average magnitude of the same parameters in the model finetuned for Context C2 is $0.0091$ (shown in Figure \ref{fig:heat-map}(b) with brighter red color pixels), which is approximately 100 times higher. This indicates two crucial aspects: \textbf{(a)} these parameters contribute minimally to model inference in Context C1, indicating redundancy for C1. \textbf{(b)} same parameters’ higher magnitudes in the model personalized in another context, C2, indicates that parameters considered unimportant in one context may be crucial in another.


\par 
The critical question arises: \emph{How can we effectively retain and transfer the valuable generic information about unseen contexts (e.g., one of the unseen contexts is C2 in the above example) to the personalized models without access to the unseen contexts?} – that this paper addresses.

\section{Problem Statement}
\label{problem-statement}
Given a generic model $\MG$, the objective is to tailor a personalized model $\MP$ specifically for a user $\Ui$ utilizing the data $\Dai$ associated with available context $\Ca$, here $\theta$ represents the parameters of the model. The primary goal is to ensure that the personalized model $\MP$ performs reasonably well on $\Ui$'s data $\Dui$ derived from unseen contexts $\Cu$. Notably, there is no overlap between the data belonging to the available and unseen contexts, that is, $\Dai \cap \Dui = \phi$.

In other words, if $\MC$ represents a \emph{conventionally-finetuned} model trained for a user $\Ui$ on data $\Dai$, then, the models trained using \textcolor{violet}{CRoP}, $\MP$, must on avg. perform better on both available $\Ca$ and unseen context $\Cu$ than $\MC$.  More formally, the learning objective can be defined as:
$$\MP = \argmin_\theta \sum_{d \in \Dai} \quad \ell(\MG, d),$$
such that $\Dai \cap \Dui = \phi$ and $$\sum_{d \in \{\Dui,\Dai\}} \ell(\MP , d) < \sum_{d \in \{\Dui,\Dai\}} \ell(\MC , d),$$ that is, the loss incurred by the resulting personalized model $\MP$ on avg. across all contexts' data is less than the loss incurred by \emph{conventionally-finetuned} model $\MC$. Here, $\ell$ represents the standard cross-entropy loss.

It is important to emphasize that the above-mentioned optimization problem restricts the usage of data only to the available context $\Ca$ and has no knowledge of data from the unseen context $\Cu$. Hence, for $d \in \Dui$ (unseen context data), the information $\ell(\MP, d)$, and $\ell(\MC, d)$ is absent during the training process. 


\section{Approach}
\label{approach}

\subsection{Rationale for The \textcolor{violet}{CRoP} Approach Design}
\label{approach-1}
As previously discussed, the generic model's parameters contain generalizable information across all contexts. Addressing the problem statement requires retaining this information to the greatest extent while enabling fine-tuning for the target user. Furthermore, our investigation revealed that different parameters hold varying degrees of importance in distinct contexts. Hence, the careful selection of subsets of model parameters for personalization and generalization is pivotal for the success of the approach, for which this paper leverages the model pruning paradigm. 
\par 
Model pruning is based on the idea that neural networks include redundant parameters, and removing these parameters minimally impacts the model's performance \citep{ThiNet-ICCV17, zhu2017prune}. Consequently, pruning the fine-tuned personalized model ensures the retention of essential parameters to maintain accuracy for context $\Ca$, meaning capturing the target user-specific traits present in context $\Ca$ in a substructure of the personalized model. Consequently, the values of the pruned parameters (that are zeroed out) can be replaced with corresponding parameter values from the generic model, effectively restoring generic knowledge learned across all contexts through those parameters of the generic model. This restoration may enhance generalizability, ensuring robust performance in unseen contexts $\Cu$. The approach presented in this paper is founded on this insightful strategy.

\subsection{\textcolor{violet}{CRoP} Approach}
\label{approach-details}

Algorithm \ref{alg:main} describes the presented approach, which takes as input: the generic model $\cM_{G}$, user $\Ui's$ data $\Dai$ for available context $\Ca$, the initial value for the coefficient of regularization $\alpha$ and tolerance for pruning $\tau$; and generates the target personalized model $\MP$. Here, $\alpha$ and $\tau$ are hyperparameters whose values can be tuned for the given data and model. 

\begin{wrapfigure}{r}{0.60\textwidth}
\vspace{-6ex}
\begin{minipage}{\linewidth}
\small{
\begin{algorithm}[H]
   \captionof{algorithm}{CRoP}
   \label{alg:main}
\begin{algorithmic}[1]
   \STATE {\bfseries Input:} $\MG$: Generic Model
  $~\diamond~$ $\Dai$: User $\Ui's$ data for available context $\Ca$ $~\diamond~$ $\alpha$: coefficient of regularization  $~\diamond~$ $\tau$: tolerance for pruning
   \STATE Train the Generic model on the personal data $\Dai$
   $$\cM^{P^a_i}_{\theta'} = \argmin_\theta \sum\limits_{d\in \Dai}  \ell(\MG, d) + \alpha \|\MG\|_1 $$
   \STATE Prune redundant parameters to obtain the pruned sub-structure
   $$\cM^{P^a_i}_{\theta^\downarrow} = \text{ToleratedPrune}(\cM^{P^a_i}_{\theta'}, \tau),$$
   where $\theta^\downarrow$ signifies that a subset of model parameters are non-zero.
   \STATE Copy the parameters of generic model to the pruned parameters in the personalized pruned model, $$\cM^{P^a_i}_{\theta''} = \begin{cases}
       \cM^{P^a_i}_{\theta^\downarrow} &, \theta^\downarrow \ne 0 \\
       \MG &, \text{otherwise}\\
   \end{cases}$$
   \STATE Finetune the personalized model on the $\Dai$ using early stopping with validation accuracy in $\Ca$  $$\MP = \argmin_\theta \sum\limits_{d\in \Dai}  \ell(\cM^{P^a_i}_{\theta''}, d) + \alpha \|\cM^{P^a_i}_{\theta''}\|_1,$$
\end{algorithmic}
\end{algorithm}}
\end{minipage} 
\vspace{-4ex}
\end{wrapfigure}

\par 
The approach initiates by finetuning the generic model $\MG$ on data $\Dai$, concurrently applying $\ell_1$ regularization to penalize model parameters (line 2). This regularization encourages sparsity by specifically targeting the magnitude of redundant parameters \citep{l1-reg}. This step is followed by the pruning of redundant weights using the `\emph{ToleratedPrune}' module (line 3). The pruned weights are then replaced by the corresponding weights from the generic model $\cM_{G}$ (line 4) to restore generalizability; this hybrid model is referred to as the \emph{`Mixed Model.'} 
This step can lead to model inconsistencies resulting in performance loss \citep{yadav2023tiesmerging, daheim2024model} for some users.
To mitigate such a loss for the affected users, as a final step, the \emph{Mixed Model} undergoes fine-tuning once again on the data from the available context $\Dai$ (line 5). The detailed explanation of each of these steps is as follows:

\paragraph{Personalized Finetuning with Penalty (Algorithm \ref{alg:main} – Step $2$):}
The approach uses data $\Dai$ to finetune the generic model $\MG$. As shown in the Section \ref{motivation} (Motivation), such conventional finetuning enhances the model's accuracy within the available context $\Ca$. Nevertheless, its performance in unfamiliar contexts may get suboptimal. Notably, during the model's fine-tuning process, we apply $\ell_1$ regularization to penalize the model weights, forcing the magnitudes of redundant parameters to be close to zero \citep{l1-reg}. The regularization coefficient $\alpha$ is a trainable parameter optimized during training to minimize the overall loss. As a result, the parameters with high magnitudes carry most of the information regarding the data patterns in $\Dai$, offering two key benefits:
\begin{enumerate}[topsep=0pt,nolistsep, leftmargin=*]
    \item Minimal loss in $\Ca$ accuracy: A high fraction of parameters have close to zero, and their removal results in minimal information loss for context $\Ca$; thus, the adverse impact of pruning in context $\Ca$ is minimized.
    \item Maximal generalization: The inclusion of regularization aids \emph{ToleratedPrune} (discussed below) module in efficiently pruning a higher number of parameters, which are then replaced with weights from the generic model. This restores information from the generic model, contributing to enhanced accuracy in context $\Cu$. 
\end{enumerate}

\paragraph{ToleratedPrune Module (Algorithm \ref{alg:main} – Step $3$):} 
Algorithm \ref{alg:prune} outlines the \emph{ToleratedPrune} module, taking a model $\cM_\theta$, pruning tolerance $\tau$, and the dataset $\cD$ as inputs. Different individuals may exhibit varying degrees of deviation in their user-specific traits \citep{gong2023dapper} compared to the generic traits learned by the generic model. As a result, fine-tuning a generic model requires different amounts of training to capture user-specific traits for each individual. Additionally, the model substructure necessary for learning these traits is shaped by regularization (Algorithm \ref{alg:main} Line 2) applied during the initial fine-tuning step. The combined effects of data heterogeneity, varying degrees of learning, and regularization lead to the necessity of requiring distinct substructures for each individual. Thus, choosing a single pruning amount for all individuals might not be the optimal pruning strategy.

\begin{wrapfigure}{r}{0.40\textwidth}
    
\vspace{-2ex}
\begin{minipage}{\linewidth}
\small{
\begin{algorithm}[H]
   \captionof{algorithm}{ToleratedPrune$(\cM,\tau, \cD)$}
   \label{alg:prune}
\begin{algorithmic}[1]
   \STATE {\bfseries Input:} $\cM_\theta$: A Model
  $~\diamond~$ $\tau$: tolerance for pruning $~\diamond~$ $\cD$: data
   \STATE Pruning Amount $p = k$
   \STATE $A_o = accuracy(\cM_\theta,\cD)$
   \REPEAT
        \STATE $\cM_{\theta^\downarrow} = \cM_\theta$
        \STATE $\cM_{\theta} = Prune(\cM_\theta,p)$
        \STATE $A = accuracy(\cM_\theta,\cD)$
        \STATE Increment Pruning Amount $p = p + k'$
   \UNTIL{$A < A_o - \tau$}
    \STATE return $\cM_{\theta^\downarrow}$
   
\end{algorithmic}
\end{algorithm}}
\end{minipage}
\vspace{-2ex}
\end{wrapfigure}

To solve this, the \emph{ToleratedPrune} module iteratively computes an optimal sub-network for each individual's available context $\Ca$. It initiates with a modest pruning amount of $k$ and incrementally increases this amount by $k'$ until the model's accuracy exhibits a drop of $\tau$ percent on $\cD$. Here, $k$ and $k'$ are hyperparameters within the range of $(0,1)$. The module returns $\cM_{\theta^\downarrow}$, representing the pruned state of the model before the last pruning iteration. The symbol $\theta^\downarrow$ suggests that only a subset of this model's parameters are non-zero. This state is such that further pruning would result in a higher accuracy loss on dataset $\cD$ than the tolerable amount $\tau$. This module performs pruning leveraging the conventional magnitude-based unstructured pruning \citep{zhu2017prune}.

Thus, step 3 in Algorithm \ref{alg:main} generates a pruned personalized model state $\cM^{P^a_i}_{\theta^\downarrow}$ whose prediction accuracy on context $\Ca$ is at most $\tau$ percent lower than that of the earlier state  $\cM^{P^a_i}_{\theta'}$ while using only a fraction of its original parameters.  The non-zero weights corresponding to these parameters contribute significantly to model inference for the available context $\Ca$. As a result, $\cM^{P^a_i}_{\theta^\downarrow}$ is essentially the \emph{minimal sub-structure} of the earlier state model $\cM^{P^a_i}_{\theta'}$, which is crucial for correct inference for context $\Ca$. This enables replacing a maximal number of zeroed-out parameters to incorporate information from unseen contexts using the generic model $\MG$ in the subsequent steps.

\paragraph{Generating the \emph{Mixed Model (Algorithm \ref{alg:main} – Steps $4 \& 5$)}:} 
For generating the \emph{Mixed Model}  $\cM^{P^a_i}_{\theta''}$, the zeroed out parameters in the pruned model $\cM^{P^a_i}_{\theta^\downarrow}$ are replaced by the corresponding parameters in the generic model $\MG$, enabling generic knowledge restoration.

Recent works have highlighted performance degradation while merging two trained models, attributing the loss to either parameter sign flip \citep{yadav2023tiesmerging} or change in gradient direction \citep{daheim2024model}. Both of these works show that directly merging two trained models can result in inconsistencies in the model, which may cause accuracy degradation, necessitating the need for a mechanism to address this. This is in line with our observation that despite the \emph{Mixed Model} exhibiting improved performance in the unseen context, there is a notable loss of accuracy in the available context for some users. \citet{finetuning-1} have shown that finetuning after merging different models can result in a performance boost. Thus, we include a final finetuning step in our personalization mechanism for the affected users. \citet{Goyal_2023_CVPR} suggests that fine-tuning process should mirror pre-training for effective generalization. Therefore, our final fine-tuning objective aligns with the pre-training objective used in Line 2 for optimal results.

\par 
During finetuning, the model state, including the \emph{mixed model}, with the best validation loss on the available context, is selected. We found that for some individuals, the \emph{mixed model} is chosen as the optimal model, which indicates that for some individuals, further finetuning is not required, and our approach can automatically handle that scenario.

\section{Experiments}
\label{sec:experiments}

This section displays the empirical efficacy of the presented approach. Section \ref{dataset-details} provides a detailed discussion of the four human sensing datasets and their pre-processing used in our evaluations: PERCEPT-R \citep{Benway2023-eu}, WIDAR \citep{WIDAR-dataset}, ExtraSensory \citep{Extrasensory-dataset} and a Stress-sensing dataset \citep{stress-sensing}, and also explains the corresponding model architectures as used in literature. Section \ref{evaluation-metrics} details the metrics of evaluation used in this work along with their practical relevance. This is followed by a detailed discussion of the empirical comparison of \textcolor{violet}{CRoP} with five baselines SHOT\cite{liang2020we}, Packnet\cite{packnet}, Piggyback\cite{Piggyback}, CoTTa\cite{cotta} and PTN\cite{PTN-baseline} in Section \ref{comparison-with-soa}. 
Additionally, detailed discussions about interesting patterns, such as the impact of generic model quality on personalization, etc., are also provided.

Notably, each dataset offers a varying distribution of data among different classes, necessitating different performance metrics like inference accuracy and F1 score. For each dataset, our evaluation stayed aligned with the metrics previously used in studies evaluated on these datasets, with further details available in the Appendix \ref{reproducibility}.
Additional details to support reproducibility, such as hyperparameters, links to the code, and computation resources utilized, are also provided in Appendix \ref{reproducibility}.



\subsection{Datasets and models}
\label{dataset-details}
This work employs four real-world human-sensing datasets to demonstrate the empirical efficacy of \textcolor{violet}{CRoP}, two of which are associated with health applications. First, the PERCEPT-R dataset has been used for binary classification for predicting the correctness of /\rotatebox[origin=c]{180}{r}/ sounds in automated speech therapy application \cite{benway-and-preston}. 
Additionally, we use the Stress Sensing dataset \cite{stress-sensing} collected using a psycho-physiological wrist-band, named Empatica E4 \citep{empatica}. To further demonstrate the efficacy of \textcolor{violet}{CRoP}, we incorporate two benchmark human-sensing datasets, which include data from the same individuals across multiple contexts: WIDAR \citep{WIDAR-dataset} and  ExtraSensory \citep{Extrasensory-dataset}. Specifically, we employ WIDAR for a 6-class classification focusing on gesture recognition using WiFi signals, and ExtraSensory for binary classification related to human activity recognition using accelerometer and gyroscope readings. Details on the datasets, preprocessing for personalized evaluations, and generic model training are discussed below.

\emph{PERCEPT-R:} The data used in this study come from corpus version v2.2.2 \cite{benway-and-preston}, which consist of single-word speech audio collected during clinical trials involving children with speech sound disorders affecting /\rotatebox[origin=c]{180}{r}/, along with age-matched peers with typical speech. The full corpus contains 179,076 labeled utterances representing 662 single-rhotic words and phrases. Each audio file is paired with a ground-truth label representing listener judgments of rhoticity, derived by averaging binary ratings (0 = derhotic, 1 = fully rhotic) from multiple listeners. 
\par 
To use this dataset for our personalization evaluation, we collaborated with clinical experts to identify and acquire annotations of $16$ participants who had correct and incorrect pronunciations of /\rotatebox[origin=c]{180}{r}/ sound at pre-treatment (baseline-phase) and during different treatment phases. 
The evaluation treats pre-treatment data as available context data, while data from other treatment phases serve as unseen context data. 

\emph{WIDAR:} WIDAR is a gesture recognition dataset collected using off-the-shelf WiFi links (one transmitter and at least three receivers). 17 users performed 15 gestures across different rooms and torso orientations. The channel state information is collected from these devices with amplitude noises and phase offsets removed as a preprocessing step. The two contexts used for the current work are decided based on the orientations of the torso data and room ID. 
Room 1 is a classroom with a number of objects (e.g., desks and chairs) in it, and Room 2 is a nearly empty hallway. The dissimilar data distributions can be attributed to the differences in the amount of interruptions in WiFi signals. We followed the same normalization methods as \citet{Z-score}.

\emph{ExtraSensory:} ExtraSensory is a human activity recognition dataset collected using the ExtraSensory mobile application. 
A number of features were collected from different cellular devices and smartwatches, though we just used the accelerometer and gyroscope features along with self-reported labels. The contexts are decided based on the location of the phone: hand, pocket, and bag.

\emph{Stress Sensing Dataset:}\label{Stress-dataset} 
This dataset measures the physiological impacts of various kinds of Stress. The dataset is collected using Empatica E4 Wristband to extract features such as EDA (Electrodermal Activity), a skin temperature sensor ($4$ Hz), etc, contributing to a total of 34 features. 
The data was collected from 30 participants with different demographics, and was labelled `Stressed' or `Calm' based on the stress-inducing or calming tasks.
For this work, we collaborated with the authors to identify participants wearing wristbands on both hands for personalization. Activities were annotated as still or moving based on movement patterns, and these annotations will be shared publicly. Contexts are defined by the wristband hand and movement state during data collection.

\subsubsection{Pre-Processing of the Datasets} 
\label{dataset-preprocessing}
We partitioned each dataset into two disjoint sets of users: (1) a generic dataset for training a generic model and (2) a personalized dataset for training a personalized model for each user. To demonstrate the context-wise robustness, we further partitioned each user's (belonging to the later set) personalized dataset into different contexts (i.e., available $\Ca$ and unseen context $\Cu$).
Table \ref{tab:my_label} presents the details of this partitioning. For PRECEPT-R, we consider data from the pre-treatment phase as the available context, and the treatment phases, where participants undergo clinical interventions, are considered the unseen context. For the Stress Sensing dataset, the context is determined by two factors: the hand on which the sensor (Empatica E4 wristband \cite{empatica}) was worn during data collection and the movement status of the individual. For WIDAR, context is determined by the room and torso orientation during data collection, while for the ExtraSensory dataset, phone's location on the user's body (e.g., hand, pocket, bag) defines the context. The term `Scenario' refers to the combination of available $\Ca$ and unseen $\Cu$ contexts as outlined in Table \ref{tab:my_label}. All datasets, along with context-wise annotations, will be made public.

\begin{table*}[t]
    \centering
    \resizebox{0.99\linewidth}{!}{
    \begin{tabular}{|c|c|c|c|c|}
    \hline
         Dataset $\rightarrow$& PERCEPT-R & WIDAR& ExtraSonsory & Stress Sensing\\
         \hline
         Total users&515&17&60&30\\
         \hline
         Users' ID for& 17,25,28,336,344,361,362,55,& 0,1,2 & 80,9D,B7,61,7C & 1,2,3\\
         Personalization&586,587,589,590,591,61,67,80 &&&\\
         \hline
         Scenario 1 &$\Ca$: Baseline Study& $\Ca$: Room-1, Torso Orientation- 1,2,3 & $\Ca$: Hand, Pocket & $\Ca$: Left hand, Still\\
         & $\Cu$: Treatment Phase &$\Cu$: Room 2, Torso Orientation- 4,5 & $\Cu$: Bag & $\Cu^1$: Right hand, Still; $\Cu^2$: Right hand, Moving\\
         \hline
         Scenario 2 & -N/A-&$\Ca$: Room 2, Torso Orientation- 4,5 & $\Ca$: Bag, Pocket & $\Ca$: Right hand, Moving\\
         && $\Cu$: Room-1, Torso Orientation- 1,2,3 & $\Cu$: Hand& $\Cu^1$: Left hand, Moving; $\Cu^2$: Left hand, Still\\
         \hline
    \end{tabular}}
    \caption{Details of data used for personalization}
    \vskip -6ex
    \label{tab:my_label}
\end{table*}

\emph{Notably, throughout the training of personalized models, \textcolor{violet}{CRoP} refrains from utilizing any information from the unseen context $\Cu$. Therefore, while the empirical study indicates an enhancement in the model's performance for one or a few unseen contexts, it is a proxy for all unseen contexts. This means that it is reasonable to anticipate a favorable performance in other unseen contexts that are not available on the dataset.} 

Notably, the stress sensing dataset has been evaluated across two different unseen context variations. In $\Cu^1$, there was one change in context—a change in hand while keeping the same movement pattern as $\Ca$. In $\Cu^2$, there were two changes—a change in both hand and movement patterns.


\subsubsection{Data segmentation for Personalization and Generic Model Training} 
\label{model-details}

\hfill
\par\emph{PERCEPT-R}: For the identified 16 participants for our personalization evaluation, their
speech data were collected longitudinally, meaning their data could be separated into available and unseen contexts versus other speakers in the corpus who only had speech data available from a one-time point. The generic model is trained for the remaining 499 participants (having /\rotatebox[origin=c]{180}{r}/ sounds disorder) using person-disjoint validation and test sets. The aim of this dataset is to identify the correctness of /\rotatebox[origin=c]{180}{r}/ sounds. 


\emph{WIDAR:} We chose 3 users for personalization since these were the only users whose data was collected in both rooms. Since the number of users in WIDAR is very small, the exclusion of all the 3 users for training the generic model would have resulted in substandard models. So, For each user, we generated different generic models by using data from the other 16 users with a 14/2 person disjoint random split for the train and validation set.  Our classification target was the 6 gesture classes: 0,1,2,3,5 and 8, corresponding to push, sweep, clap, slide, draw a circle, and draw zigzag, respectively, as evaluated in the original work \cite{WIDAR-dataset}. 


\emph{ExtraSensory:} We chose 5 users for personalization, and the generic model is trained on 42 users, with 10 users being left out for person-disjoint validation. The two target classes are walking and sitting.  


\emph{Stress Sensing:} For this binary task, three users with data across all contexts were selected for personalization. The generic model was trained on data from 21 users, with six users for disjoint validation and test sets.

\emph{Generic model training:} We employ the model architectures and training methods used in the literature corresponding to each dataset: 
PERCEPT-R \citep{Benway2023-eu}, WIDAR \citep{WIDAR-dataset}, ExtraSensory \cite{metasense,invariant_feature_learning}, Stress-sensing\citep{eren2022stress}. Details are provided in Appendix \ref{model-architectures}.


\subsection{Metrics for evaluation}\label{evaluation-metrics}

To establish the efficacy of \textcolor{violet}{CRoP}, we quantify the extent of \emph{personalization} and \emph{generalization} achieved through the presented approach. Personalization is gauged by comparing our model's $\MP$ accuracy relative to the generic model $\MG$, while for generalization, we assess the accuracy of our model $\MP$ against \emph{conventionally-finetuned} personalized models $\MC$. 
\par 
\citet{metrics} argued that directly comparing model accuracies under distribution shifts is not ideal. They introduced `effective robustness,' a metric that assesses performance relative to an accuracy baseline. Since we aim to compare our models against two baselines — the generic and conventionally finetuned models — we adopt the `effective robustness metric' and introduce two specific metrics for our comparison, detailed below. Both of these metrics consider classification accuracy in the available $\Ca$ and unseen $\Cu$ contexts.

If $\cA(\cM,\cD)$ represents the classification accuracy of the model $\cM$ for dataset $\cD$ and $n$ is the number of users selected for personalization, the metrics of evaluations can be described as follows :

\begin{enumerate}[topsep=0pt,nolistsep, leftmargin=*]
    \item Personalization $(\Delta_P)$: It is defined as the sum of the difference between the accuracy of $\MP$ and $\MG$ over all the contexts averaged over all users
    $$
        \Delta_P =  \frac{1}{n}\sum_{\Ui} \sum_{\cC \in \{\Ca,\Cu\}} (\cA(\MP,\cC) - \cA(\MG,\cC)) $$
    \item Generalization $(\Delta_G)$: It is defined as the sum of the difference between the accuracy of $\MP$ and $\MC$ over all the contexts averaged over all users.
    $$
        \Delta_G =  \frac{1}{n}\sum_{\Ui} \sum_{\cC \in \{\Ca,\Cu\}} (\cA(\MP,\cC) - \cA(\MC,\cC)) $$
    
\end{enumerate}

\emph{In summary,} the metric $\Delta_P$ suggests how well the model $\MP$ performs as compared to the generic model $\MG$. Since the generic model has not learned the user-specific patterns. This metric quantifies \textcolor{violet}{CRoP}'s ability to learn user-specific patterns. On the other hand, conventionally finetuned models $\MC$ may learn user-specific patterns and forget the generic information of different contexts. Thus, the metric $\Delta_G$ quantifies \textcolor{violet}{CRoP}'s ability to retain generic information.

All the results in this section are computed as an average of accuracy obtained for three random seeds.

\subsection{Comparison with State-of-the-art}
\label{comparison-with-soa}
To demonstrate the efficacy of \textcolor{violet}{CRoP} in achieving personalization $\Delta_P$ while maintaining generalization $\Delta_G$, we compare \textcolor{violet}{CRoP} with 5 state-of-the-art approaches SHOT \citep{liang2020we}, PackNet \cite{packnet}, Piggyback \cite{Piggyback}, CoTTA \cite{cotta}, and PTN \cite{PTN-baseline}. 

\begin{table*}[t]
    \centering
    \resizebox{0.99\textwidth}{!}{
    \begin{tabular}{|c|c|lr|lr|lr|lr|lr|lr|}
    \hline
         & Approach & \multicolumn{2}{|c|}{SHOT} &
         \multicolumn{2}{|c|}{Packnet} &\multicolumn{2}{|c|}{Piggyback} &\multicolumn{2}{|c|}{CoTTA} &\multicolumn{2}{|c|}{PTN} &\multicolumn{2}{|c|}{\textcolor{violet}{CRoP}}  \\
        \hline
         Dataset&Scenrio& $\Delta_P$& $\Delta_G$& $\Delta_P$& $\Delta_G$&
         $\Delta_P$& $\Delta_G$&
         $\Delta_P$& $\Delta_G$&
         $\Delta_P$& $\Delta_G$&
         $\Delta_P$& $\Delta_G$\\
         \hline
         PERCEPT-R & Scenario 1&-3.11 &-5.62&0.10&-2.41&-25.31&-27.83&-45.06&-47.58&-1.16&-3.68&\textbf{5.08}&\textbf{2.57}\\
         \hline
         Stress Sensing & Scenario 1 &-8.19&-62.16&54.70&0.70&43.89&-10.12&21.93&-32.07&1.23&-52.76&\textbf{67.81}&\textbf{13.81}\\
        Single context Change& Scenario 2 &8.90&-63.27&75.80&3.64&66.22&-5.94&51.47&-20.69&21.76&-50.40&\textbf{85.25}&\textbf{13.08}\\
        \hline
        Stress Sensing & Scenario 1 &-0.49&-47.24&52.46&10.08&32.40&-9.97&30.59&-11.78&6.40&-35.98&\textbf{54.38}&\textbf{12.00}\\
        Double context Change& Scenario 2 &3.57&-45.49&41.68&-7.36&42.76&-6.25&33.85&-15.19&12.30&-36.74&\textbf{59.21}&\textbf{10.15}\\
        \hline
        WIDAR & Scenario 1 &1.67&-0.48&-0.24&-2.37&0.84&-1.28&-1.05&-3.18&-1.99&-4.37&\textbf{8.56}&\textbf{6.43}\\
        & Scenario 2 &1.28&-0.03&-3.55&-5.16&-8.97&-10.57&1.81&0.21&0.00&-2.85&\textbf{5.90}&\textbf{4.30}\\
        \hline
        ExtraSensory & Scenario 1 &7.63&-10.31&12.19&-5.76&5.03&-12.91&-0.6&-18.54&1.69&-16.72&\textbf{17.49}&\textbf{-0.46}\\
        & Scenario 2 &8.17&2.99&1.33&-3.85&5.22&0.04&-3.43&-8.62&-3.02&-7.47&\textbf{13.52}&\textbf{8.17}\\
        \hline
            \end{tabular}}
    \caption{Comparison of \textcolor{violet}{CRoP} with baseline approaches under the metrics of Personalization ($\Delta_P$) and Generalization ($\Delta_G$).} 
    \label{tab:main-table}
    \vskip -4ex
\end{table*}

Table \ref{tab:main-table} compares the performance of \textcolor{violet}{CRoP} with aforementioned baseline approaches. The values for $\Delta_P$ and $\Delta_G$ are computed as average over all the participants used for personalization for each dataset. The detailed results for participant-specific evaluations for each dataset are provided in  Appendix \ref{appendix:detailed-results} and  Appendix \ref{apendix:error-bars} shows the errors bars for our approach.  

Table \ref{tab:main-table} shows that \textcolor{violet}{CRoP} significantly outperforms all state-of-the-art (SOTA) methods. 
On average, the personalization benefits $\Delta_P$ achieved by SHOT, PackNet, Piggyback and CoTTA are $2.16$, $26.05$, $18.01$, $9.95$, and $4.13$  percent points, respectively, while \textcolor{violet}{CRoP} can achieve $35.23$ percent points. However, while comparing $\Delta_G$, one can observe that personalized training using SHOT, PackNet, Piggyback and CoTTA harms generalizability by $-25.73$, $-1.39$, $-9.43$, $-17.49$ and $-23.44$ percent points respectively. On the other hand, \textcolor{violet}{CRoP} shows an average generalization benefit of $7.78$.  %

Notably, domain adaptation and continual learning methods, such as SHOT, Packnet, Piggyback, CoTTA, and PTN, leverage data from new contexts to adapt models while preserving knowledge from previous contexts or tasks. However, the problem addressed in this work requires the model to perform reasonably well in a completely unseen context, for which no data—labeled or unlabeled—is available during training. As a result, all baseline approaches struggle significantly in this scenario. Since the metrics in Section \ref{evaluation-metrics} evaluate performance across both available and unseen contexts, the substantial performance drop in the unseen context negatively impacts overall results. 
Furthermore, the unsupervised approaches SHOT, CoTTA, and PTN achieved lower $\Delta_P$ and $\Delta_G$ than supervised approaches Packnet, Piggyback, and \textcolor{violet}{CRoP}, which is in line with literature \cite{supvsunsup}.

The following discusses notable patterns observed in this section's evaluations in Table \ref{tab:main-table}, insights into the characteristics of static personalization across various human-sensing applications.

\subsubsection{For stress-sensing dataset, all the unsupervised approaches show better performance on double context change than the single-context change}
For Scenario 1, $\Ca$ contains samples of data collected using the left hand while staying still. For $\Cu^1$ (single context change), the hand changes to the right while staying still, but in $\Cu^2$ (double context change), both hand and motion status changes. Since $\Cu^2$ incurs greater shift in context than $\Cu^1$, one can expect that a model trained on $\Ca$ will show similar or worse results for $\Cu^2$ as compared to $\Cu^1$. However, unsupervised approaches—SHOT, CoTTA, and PTN—exhibit higher $\Delta_P$ and $\Delta_G$ values under $\Cu^2$ than $\Cu^1$.

\begin{wrapfigure}{r}{0.35\textwidth}
    \centering
    \subfloat[Perspective - models]{\includegraphics[width=\linewidth]{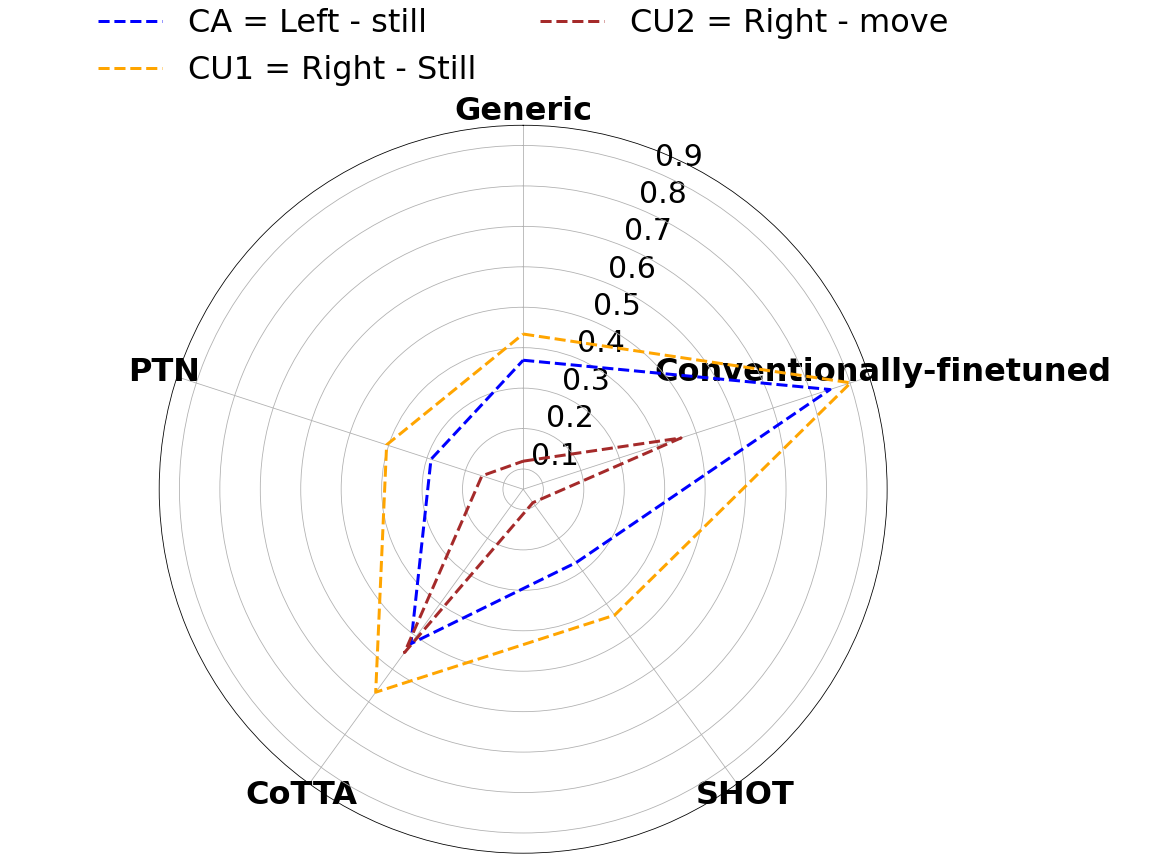}}
    \label{fig:stress-sensing-user3-details-models}
    
    \subfloat[Perspective - Contexts]{\includegraphics[width=\linewidth]{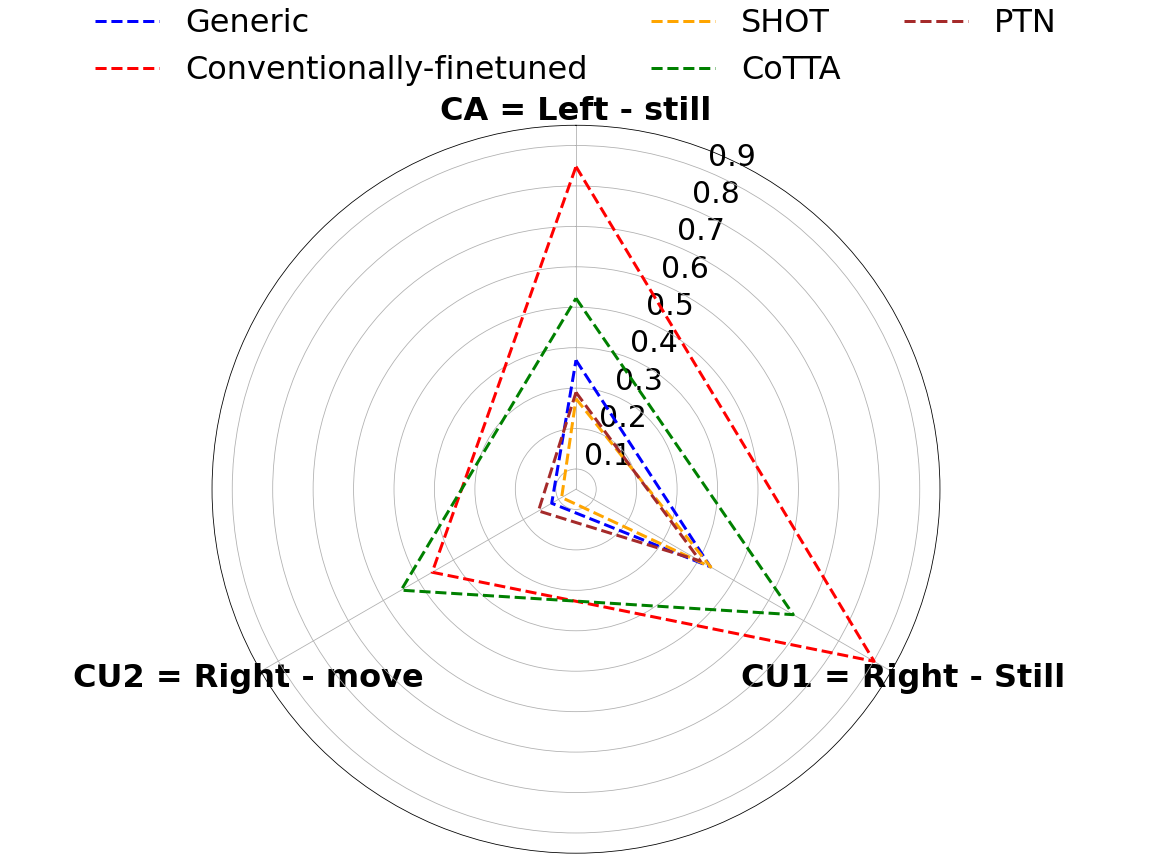}}
    \label{fig:stress-sensing-user3-details-contexts}
    \caption{Detailed results for Stress sensing dataset User 3 (Scenario 1)  through F1 score}
    \label{fig:stress-sensing-user3-details}
    \vspace{-4ex}
\end{wrapfigure}

User-specific results, however, indicate that models actually perform worse on $\Cu^2$.
Figure \ref{fig:stress-sensing-user3-details}(a) shows detailed results in terms of F1 score for User 3 for generic model $\MG$, conventionally-finetuned model $\MP$, and models generated using unsupervised approaches: SHOT, CoTTA and PTN under Scenario 1. 
As expected, all models perform worse with the double context change $\Cu^2$ (CU2 = Right-move) compared to the single context change $\Cu^1$ (CU1 = Right-still). Despite this, the $\Delta_P$ and $\Delta_G$ values suggest otherwise, explained as follows:



First, consider the $\Delta_P$ results, with the generic model as the accuracy baseline.
Figure \ref{fig:stress-sensing-user3-details}(b) shows that the generic model’s F1 score drops significantly with the double context change $\Cu^2$ compared to the single context change, $\Cu^1$, depicted on the blue lines. 
Although the performance benefits using the three approaches are almost similar in both unseen contexts, the difference in the performance as compared to the generic model is higher for double context change, thus resulting in higher $\Delta_P$
Similarly, for $\Delta_G$, with the conventionally finetuned model as the baseline, the best results are seen in the available context, $\Ca$ (Left-still). Its performance is only slightly affected by the single context change but drops significantly in $\Cu^2$ due to movement during data collection, resulting in a higher $\Delta_G$ for the double context change.


Similar patterns were observed in other users, indicating that the poor performance of $\MG$ and $\MP$ during double context change drives the higher $\Delta_P$ and $\Delta_G$ values, giving an illusion of better results for double context change scenarios.

\subsubsection{Significant performance drop for PERCEPT-R using SOTA approaches}

For the Percept-R dataset, personalization applied to the 16 participants reveals that not all participants benefit equally. In some cases, the global models already perform well, leaving little to no room for improvement through personalization. In such cases, performance can even drop due to overfitting on the available context data. Our approach mitigates this by selecting the best model based on the highest validation accuracy from the available context, thus avoiding overfitting. Furthermore, CRoP introduces a trainable regularization coefficient during the initial finetuning phase, which penalizes model parameters while minimizing classification error (Algorithm \ref{alg:main} line 2 - trainable parameter $\alpha$). This enables the maximal removal of less important weights during the ToleratedPrune step without impacting performance in the available context (Algorithm \ref{alg:main} line 3). Additionally, we found that finetuning after the model mixing step was often unnecessary, as the mixed model typically emerged as the best choice (Algorithm \ref{alg:main} line 5). In contrast, the baseline approaches lack these constraints, leading to performance degradation in both the available and unseen contexts, negatively affecting the overall results.

\subsubsection{Inferior generic model performance leads to higher gain in personalization} \label{cross-context} For  ExtraSensory and Stress-sensing datasets, certain users experience sub-optimal performance with the generic model $\MG$ and the personalized finetuning helped not only in available context but also in the unseen contexts. For instance, the generic model showed suboptimal performance for certain users such as user `80' for ExtraSensory dataset under Scenario 1 and user `3'  for Stress-sensing dataset. Such users show higher benefit even in the unseen context when personalized using available context data, highlighting the importance of user-specific patterns.

\subsubsection{Significant Performance gains for the stress-sensing dataset for all approaches} Psychophysiological stress response is inherently heterogeneous in inter- and intra-user scenarios \cite{nagaraj2023dissecting}, leading to subpar performance of the generic model without personalization. As discussed in Section \ref{cross-context}, subpar performance of generic models results in higher gain during personalization. This is evident from the significantly high $\Delta_P$ values achieved by most approaches for the stress-sensing dataset in Table \ref{tab:main-table}. 


These evaluations confirm that models personalized with \textcolor{violet}{CRoP} exhibit higher generalizability to unseen contexts, making them more intra-user robust.

\section{Empirical Justification for CRoP}
\label{justification}

This section empirically justifies and discusses how the use of different components of \textcolor{violet}{CRoP} helps in incorporating personalization ($\Delta_P$) and generalization ($\Delta_G$).





\begin{table}[]
    \centering
    \resizebox{0.55\textwidth}{!}{
    \begin{tabular}{|c|c|c|c|}
    \hline
         Model State& User 0 & User 1 & User 2\\ 
         \hline
         $\MG$ - Generic &25& -389 & 12416\\
         $\cM^{P^a_i}_{\theta'}$ - Initial finetuned model (Alg1. Line 2) &-973& - 1204& -5131\\
         $\cM^{P^a_i}_{\theta^\downarrow}$ - Pruned (Alg1. Line 3)  &3777& 7002& -4197\\
         $\cM^{P^a_i}_{\theta''}$ - Mixed model (Alg1. Line 4) &4534&7254& 8610\\
         $\MP$  - Final model (Alg1. Line 5) &4534&7254&  -2541\\
         
         \hline
         \multicolumn{4}{|c|}{Generalizability introduced by the model mixing (Alg1. Line 5) step}\\
         \hline
         $\cM^{P^a_i}_{\theta'} - \cM^{P^a_i}_{\theta^\downarrow}$ - Complementary to pruned model &-285&-95& -2409 \\ Corresponding Parameters in generic model &160&53& 10297\\

         \hline
    \end{tabular}}
    \caption{GIP for different model states for 3 users chosen for personalization for WIDAR Dataset.}
    \label{tab:new_gip}

    \centering
    \resizebox{0.99\linewidth}{!}{
    \begin{tabular}{|c|cc|cc|cc|cc|cc|cc|cc|c|}
        \hline
        Model& \multicolumn{2}{|c|}{$\MG$} & \multicolumn{2}{|c|}{$\cM^{P^a_i}_{\theta'}$} & \multicolumn{2}{|c|}{$\cM^{P^a_i}_{\theta''}$} &\multicolumn{2}{|c|}{$\cA(\cM^{P^a_i}_{\theta''},\cC) - \cA(\cM^{P^a_i}_{\theta'},\cC)$} &\multicolumn{2}{|c|}{$\MP$}&\multicolumn{2}{|c|}{$\cA(\MP, \cC) -\cA(\cM^{P^a_i}_{\theta'},\cC) $} &\multicolumn{2}{|c|}{$\cA(\MP, \cC) -\cA(\MG,\cC) $}& Pruning\\
        & \multicolumn{2}{|c|}{Generic Model}& \multicolumn{2}{|c|}{Initial finetuning} & \multicolumn{2}{|c|}{Mixed model} & \multicolumn{2}{|c|}{Effect of mixing} & \multicolumn{2}{|c|}{Final Model} & \multicolumn{2}{|c|}{Final Model vs Initial Finetuning} & \multicolumn{2}{|c|}{Final Model vs Generic Model} & amount\\
        \hline
         User& $\Ca$ & $\Cu$ & $\Ca$ & $\Cu$ &  $\Ca$ & $\Cu$ & $\Ca$ & $\Cu$ &$\Ca$ & $\Cu$&$\Ca$ & $\Cu$ &$\Ca$ & $\Cu$ & \%age\\
         \hline
         0&64.44 & 76.84& 82.44&68.44& 84.02&70.14&\textcolor{cyan}{+1.58}&\textcolor{cyan}{+1.70}&84.02&70.14 & \textcolor{cyan}{+1.58}&\textcolor{cyan}{+1.70}& \textcolor{cyan}{+19.58}&\textcolor{orange}{-6.7}&40\\
         1&61.61 & 77.79&86.67&47.77&86.49&54.45&\textcolor{orange}{-0.18}&\textcolor{cyan}{+6.68} & 86.49&54.45 &\textcolor{orange}{-0.18}&\textcolor{cyan}{+6.68}& \textcolor{cyan}{+24.88}&\textcolor{orange}{-23.34}&50\\
         2&43.36 & 79.88&71.28&58.20&70.70&63.04&\textcolor{orange}{-0.58}&\textcolor{cyan}{+4.84}& 77.14&59.37 &\textcolor{cyan}{+5.86}&\textcolor{cyan}{+1.17} & \textcolor{cyan}{+33.78} &\textcolor{orange}{-20.51}&85\\
         \hline
         Average&&&&&&&\textcolor{cyan}{+0.27}& \textcolor{cyan}{+4.07}&&&\textcolor{cyan}{+2.42}&\textcolor{cyan}{+3.18}& \textcolor{cyan}{+26.08}&\textcolor{orange}{-16.85}&\\
         \hline
    \end{tabular}}
    \caption{Performance Comparison of model states after initial finetuning (Algorithm \ref{alg:main} line 2) $\cM^{P^a_i}_{\theta'}$ , model mixing (Algorithm \ref{alg:main} line 4) $\cM^{P^a_i}_{\theta''}$ and final finetuning (Algorithm \ref{alg:main} line 5) $\MP$ for WIDAR dataset under Scenario 1.Effects of model mixing are computed by comparing Mixed model with initial finetuned model.}
    \label{tab:before-and-after-mixing}

    \centering
    \resizebox{0.99\linewidth}{!}{
    \begin{tabular}{|c|c|c|c|c||c|c|c|c||c|c|c|c|}
    \hline
         Euclidean Distance& \multicolumn{4}{c||}{User0} & \multicolumn{4}{c||}{User1} &\multicolumn{4}{c|}{User2} \\
         \hline
         &Initial Finetuned & Pruned & Mixed & Final  & Initial Finetuned & Pruned  & Mixed  & Final &Initial Finetuned  & Pruned  & Mixed  & Final  \\
         \hline
         Generic model& 10.05 & 13.09& 1.62&1.62& 10.31 & 17.73 &2.16 & 2.16 & 19.85 & 61.26 & 4.36 & 15.18 \\
         Initial Finetuned model &-&2.14&8.92&8.92 & - & 5.50 & 8.84 & 8.84 & - & 15.76 & 16.19 & 4.60\\
         Pruned Model &-&-&11.47& 11.47&  - & - & 15.57 & 15.57 & - & - & 56.91 & 31.46 \\
         Mixed Model &-&-&-&0.0& - & - & - & 0.0 & - & - & - & 6.48\\
         \hline
    \end{tabular}}
    \caption{Euclidean distance between different model states for three users chosen for personalization for WIDAR dataset}
    \label{tab:distance}
    \vspace{-4ex}
\end{table}

\subsection{How different steps of \textcolor{violet}{CRoP} facilitate generalizability}
\label{sec:empirical-justification}
This section empirically discusses how each step of \textcolor{violet}{CRoP} (Algorithm \ref{alg:main}) facilitates intra-user generalizability, enhancing similarity in the model's behavior towards available $\Ca$ (available during personalization finetuning) and unseen contexts $\Cu$. The following sections outline the metrics, evaluation setup, and detailed analysis.

\paragraph{\textbf{Metric for generalizability:}} \citet{shi2021gradient} introduced the use of gradient inner product (GIP) to estimate the similarity between a model's behavior across different domains. 
If the model incurs gradients $G_i$ and $G_j$ for two domains $D_i$ and $D_j$ respectively, the GIP $= G_i * G_j$ is estimated as $(\|G_i + G_j\|_2)^2 - (\|G_i\|_2^2 + \|G_j\|_2^2)$, where $\|.\|_2$ signifies $\ell_2$-norm. The gradients $G_i$ and $G_j$ for the same model are computed using the general cross-entropy loss with a single backward pass on the samples corresponding to $D_i$ and $D_j$. The dimension of resulting gradients is the same as that of the model state. Through $\ell_2$-norm, the GIP computation considers gradients accumulated over all the parameters. 


The sign of GIP, i.e., $G_i * G_j$, indicates how the model treats the two domains:

\begin{itemize}[topsep=0pt,nolistsep]
    \item $G_i * G_j > 0$ : signifies that the gradient for both domains has the same direction, which means that if a model takes a gradient step in the direction of $G_i$ or $G_j$, its performance across both domains $D_i$ and $D_j$ will improve. In essence, it reflects a higher generalizability.

    \item $G_i * G_j < 0$ : suggests the opposite directions of the gradients; thus, improving the model's performance of one domain will worsen the performance of the other. This reflects lower generalizability.
\end{itemize}
\emph{\citet{shi2021gradient} also showed that higher GIP values with positive sign in two domains indicate more consistent behavior between the domains.} Thus, we use GIP to quantify intra-user generalizability. Specifically, we treat the contexts $\Ca$ and $\Cu$ as two distinct domains, $D_a$ and $D_u$, and compute the gradients $G_a$ and $G_u$ for a given model state to calculate the GIP value.

\paragraph{\textbf{Evaluation setup}} This section's evaluations consider the three users chosen for personalization for the WIDAR dataset under Scenario 1 which constitutes two non-overlapping contexts - the available context $\Ca$: Room-1, Torso Orientation- 1,2,3  and the unseen context $\Cu$: Room-1, Torso Orientation - 1,2,3 (as shown in Table \ref{tab:my_label}). GIP is computed considering these contexts as distinct domains. While the previous sections presented results averaged across multiple seeds, this section presents results from a single random seed ($369284$) to provide a clear comparison of the model states obtained at each step of the algorithm.


\paragraph{\textbf{Empirical result discussion}} Table \ref{tab:new_gip} shows the GIP of the generic model and the different states of the personalized model achieved after each step of Algorithm \ref{alg:main}. To further support the discussion, Table \ref{tab:before-and-after-mixing} shows the inference accuracy for contexts $\Ca$ and $\Cu$ of these different model states along with the pruning amount incurred during the \emph{ToleratedPrune} step.
Additionally, Table \ref{tab:distance} shows the Euclidean distances between different model states for a user, highlighting the parameter changes introduced at each step of Algorithm \ref{alg:main}. According to the Tables \ref{tab:new_gip}, \ref{tab:before-and-after-mixing}, and \ref{tab:distance} results, a detailed analysis of the impact of each algorithm step is provided below.

\begin{itemize} [leftmargin=*]
    \item \emph{Initial Finetuning (Algorithm \ref{alg:main} Line 2):} This step aims to learn user-specific traits using limited context $\Ca$ data. While the accuracy of the initial finetuned model for context $\Ca$ improves as compared to the generic model, it drops significantly for unseen context $\Cu$. This pattern is consistent across all users, as shown in Table \ref{tab:before-and-after-mixing}. Thus, initial finetuning results in generalizability loss which is supported by the drop in GIP values as shown in Table \ref{tab:new_gip}. Notably, User 2 experiences a much larger drop in GIP compared to Users 0 and 1. Table \ref{tab:before-and-after-mixing} shows that the generic model performs worse on User 2's $\Ca$ compared to the others, suggesting that User 2's user-specific data patterns differ significantly from the generic patterns. Thus, User 2 requires greater parameter adjustments (i.e., larger gradient updates) as evidenced by a higher Euclidean distance of $19.85$ units between the generic and finetuned model as compared to approximately $10$ units for others (Table \ref{tab:distance}), leading to a higher difference in GIP between the generic and finetuned models.

    \item \emph{Tolerated Prune (Algorithm \ref{alg:main} Line 3):} This module aims to find a maximal pruning amount for the personalized model without losing performance in the context $\Ca$. These pruning amounts vary among different users. For instance, User 0 and 1 undergo $40\%$ and $50\%$ pruning, respectively, while User 2 undergoes an extensive pruning amount of $85\%$ (shown in Table \ref{tab:before-and-after-mixing}). This is so because the training loss contains a regularization component (Algorithm \ref{alg:main} Line 2), User 2 undergoes higher parameter adjustments, i.e., larger gradient updates, which in turn results in higher regularization. Thus, the important weights learned for User 2 are restricted to a smaller subnetwork, which contributes to approximately $15\%$ parameters, thus allowing higher pruning. This is further supported by the higher Euclidean distance between the initial-finetuned and pruned model for User 2 as compared to other users (Table \ref{tab:distance}).

    The effect of the regularizer on the variable pruning amount highlights a key aspect of Algorithm \ref{alg:main}. For individuals requiring larger parameter adjustments (through larger gradient updates) —indicated by significant deviations of model parameters from the generic model—the risk of losing generalizability increases. In such cases, the regularizer ensures higher polarization of parameters allowing higher pruning, restoring more parameters from the generic model during the subsequent mixing steps, which we will demonstrate for User 2 in the following discussion. This facilitates greater recovery, enhancing generalizability.
    
   Additionally,  model pruning has been shown to enhance generalization in  literature\citep{jin2022prunings}, and we also observed an increase in GIP value in the pruned model (Step $3$) as shown in Table \ref{tab:new_gip}. On further analyzing the sub-network of parameters selected for removal during the pruning step ($\cM^{P^a_i}_{\theta'} - \cM^{P^a_i}_{\theta^\downarrow}$, complementary to the pruned model), we observed that the GIP corresponding to this subnetwork was negative (in Table \ref{tab:new_gip}), contributing to discrepancy in model behavior, and removal of these parameters improves generalizability.
    
    \item \emph{Model Mixing (Algorithm \ref{alg:main} Line 4):} In this step, the parameters removed during the previous step are replaced by corresponding parameters from the generic model. Table \ref{tab:new_gip} shows that this step further improves GIP. The details for this increase are discussed below.
    
    The parameters removed in the previous step had negative GIP, while the corresponding parameters in the generic model alone exhibited positive GIP 
    (Table \ref{tab:new_gip}, last row). Restoring these parameter values from the generic model enhances GIP in the mixed models, as shown in Table \ref{tab:new_gip}, improving the generalizability. Additionally, the reduced distance between the generic and mixed models (Table \ref{tab:distance}) indicates that this step aligns the personalized model closer to the more generalizable generic model.  
    Notably, for User 2, the distance reduction is significantly greater due to a larger number of parameters restored from the generic model. This restoration leads to a substantial increase in the GIP value (Table \ref{tab:new_gip}), clearly demonstrating improved generalization.

    \item \emph{Final Finetuning (Algorithm \ref{alg:main} Line $5$):} Finally, as discussed in Section \ref{approach-details}, model mixing may result in inconsistencies in the model parameters' signs and gradients which can have a degrading impact on model performance for some users. We employ the final finetuning to address that. As evident from table \ref{tab:distance}, User 0 and User 1 do not require fine-tuning, as indicated by the Euclidean distance between the final and the mixed model remaining $0$. However, User 2 does undergo final fine-tuning, as shown by a non-zero difference between the mixed and final model in Table \ref{tab:distance}, further improving the overall performance.

    
    Final fine-tuning may slightly reduce generalizability compared to the mixed model. Table \ref{tab:new_gip} supports this, showing a drop in GIP for the final fine-tuned model, leading to lower performance in $\Cu$ for User 2. However, overall performance across $\Ca$ and $\Cu$ remains higher than the mixed model. More importantly, despite this drop, the final model outperforms the initial fine-tuned model (from Line 2 in Algorithm \ref{alg:main}) in both GIP and performance on $\Ca$ and $\Cu$, indicating improved generalization and personalization.

\end{itemize}

\subsection{TSNE visualization}

\label{tsne-visualization}

\begin{figure}
    \centering
    \begin{subfigure}[b]{0.24\linewidth}
\includegraphics[width=\linewidth]{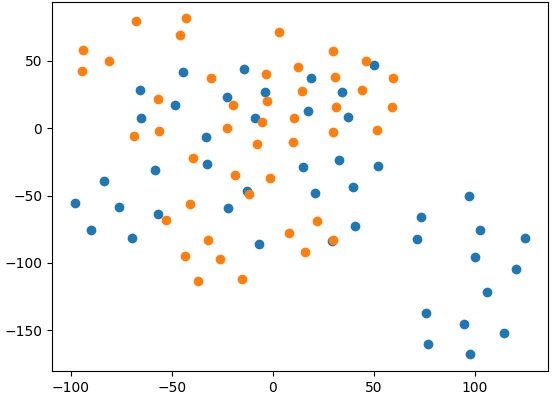}
         \caption{$\Ca$ - Generic Model\\ Accuracy = $72.99\%$}
    \end{subfigure} 
    \begin{subfigure}[b]{0.24\linewidth}\includegraphics[width=\linewidth]{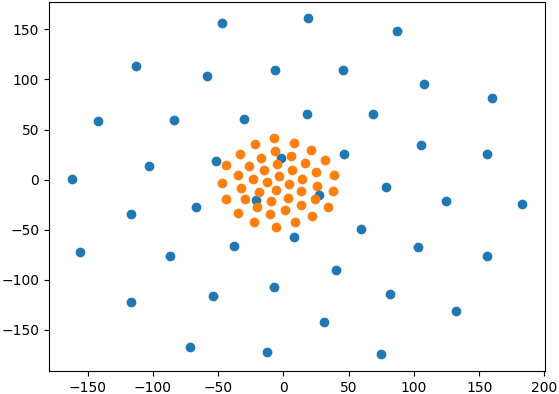}
         \caption{$\Ca$ - Initial finetuned \\ Accuracy = $78.61\%$
         }
    \end{subfigure}
    \begin{subfigure}[b]{0.24\linewidth}\includegraphics[width=\linewidth]{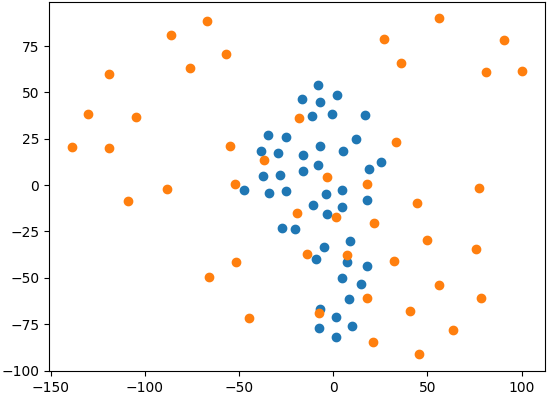}
         \caption{$\Ca$- Mixed Model \\
         Accuracy = $68.74\%$
         }
    \end{subfigure}
    \begin{subfigure}[b]{0.24\linewidth}\includegraphics[width=\linewidth]{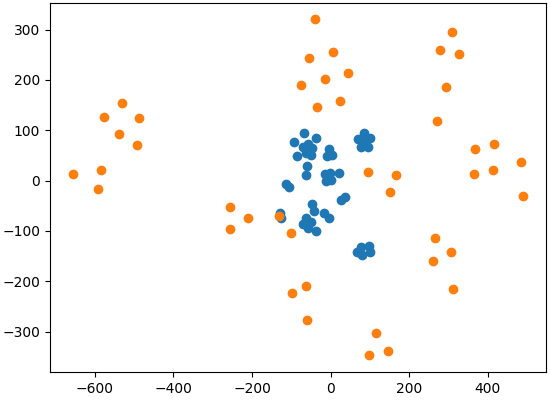}
         \caption{$\Ca$ - Final Model \\ Accuracy = $83.18\%$
         }
    \end{subfigure}

    \centering
    \begin{subfigure}[b]{0.24\linewidth}
\includegraphics[width=\linewidth]{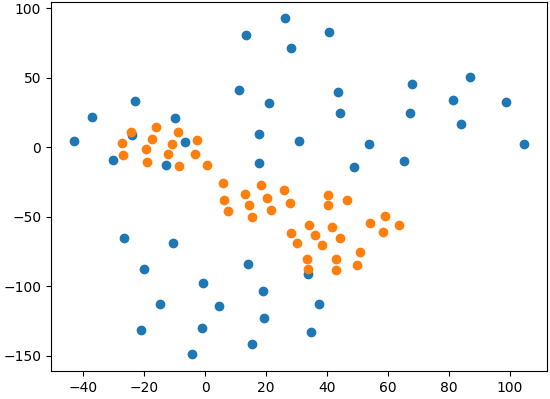}
         \caption{$\Cu$ - Generic Model \\ Accuracy = $92.32\%$}
    \end{subfigure} 
    \begin{subfigure}[b]{0.24\linewidth}\includegraphics[width=\linewidth]{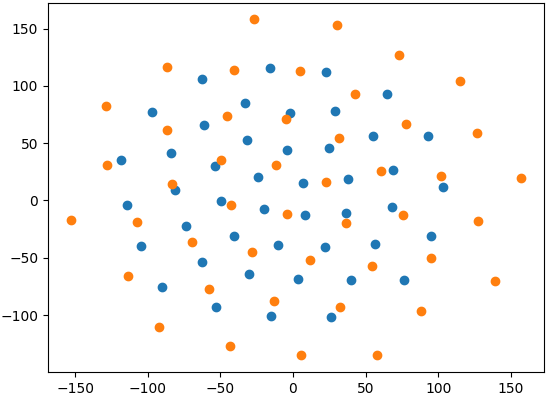}
         \caption{$\Cu$ - Initial finetuned \\ Accuracy = $67.26\%$
         }
    \end{subfigure}
    \begin{subfigure}[b]{0.24\linewidth}\includegraphics[width=\linewidth]{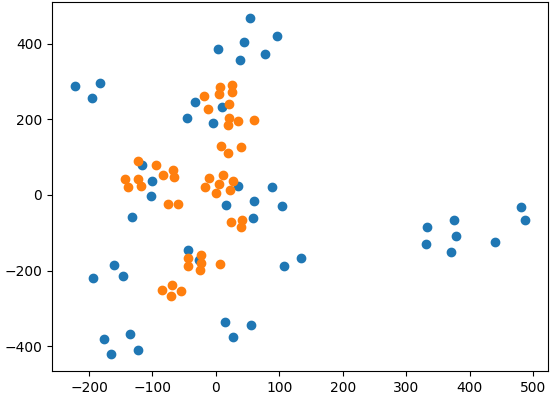}
         \caption{$\Cu$ - Mixed Model \\Accuracy = $88.62\%$
         }
    \end{subfigure}
    \begin{subfigure}[b]{0.24\linewidth}\includegraphics[width=\linewidth]{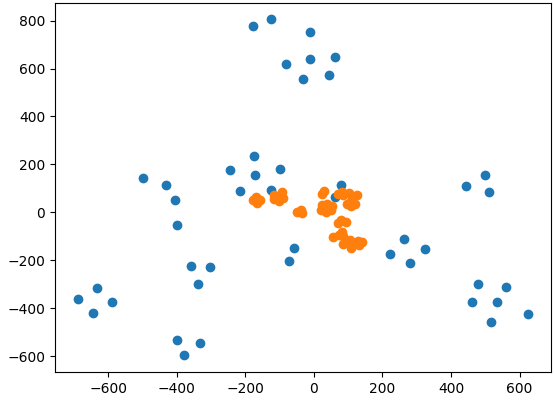}
         \caption{$\Cu$ - Final Model \\ Accuracy = $89.39\%$
         }
    \end{subfigure}
    \caption{t-SNE plots for user '7c' in ExtraSensory Dataset for Scenario 2 where avalable context $\Ca$ (`phone in pocket' and 'phone in bag') and  $\Cu$ ('phone in hand')context. The colors of the samples indicate their class membership.}
    \vskip -2ex
    \label{fig:tsne}
\end{figure}

This section further demonstrates the generalizability benefits of different steps of Algorithm \ref{alg:main} by employing t-SNE plots to visualize feature separability across classes in different model states. 
Since t-SNE represents higher dimensional features to a 2-dimensional space, we select the ExtraSensory dataset, with is a binary classification problem (walking vs sitting activity) for ease of observability. Figure \ref{fig:tsne} presents t-SNE plots for four model states—generic, initial fine-tuned, mixed, and final—showing features along with their class membership (indicated by colors). These features are extracted for both available and unseen contexts for user ‘7c,’ personalized using available context data $\Ca$ under Scenario 2, and derived from a GRU-based model trained on the ExtraSensory dataset. The model constitutes two parts: feature-extractor and classifier. The feature-extractor contains three batch-normalized 1D convolution layers followed by a linear layer that feeds into a batch-
normalized recursive (GRU) layer and two linear layers to generate embeddings. The embeddings generated by the feature extractor are used to generate the t-SNE plots.
(model details are provided in Appendix \ref{model-architectures}).

Figure \ref{fig:tsne}(a) shows that the features of different classes extracted by the generic model for the available context $\Ca$ have high overlap, leading to a low accuracy of approximately $73\%$. On the other hand, the same generic model shows better separability for unseen context $\Cu$, resulting in an accuracy of around $92\%$ (Figure \ref{fig:tsne}(e)).
As discussed in earlier sections, finetuning the generic model improves the model's performance in the available context $\Ca$, which is evident from $6\%$ gain in accuracy in $\Ca$ and better separability of features as shown in Figure \ref{fig:tsne}(b). At the same time, unseen context $\Cu$ suffers, resulting in a significant accuracy drop of around $25\%$ and an increase in the overlap between the extracted features for the two classes, as shown in Figure \ref{fig:tsne}(f). Next, the model mixing step brings generic information from the generic model to the personalized model, and as expected, an improvement in the model's performance for $\Cu$ can be seen through a gain in accuracy and separability of features (Figure \ref{fig:tsne}(g)). Simultaneously, as discussed in Section \ref{approach-details}, the inconsistencies in the mixed model result in performance degradation, especially for context $\Ca$ (Figure \ref{fig:tsne}(c)). Final finetuning aims to address this issue, and as depicted by Figure \ref{fig:tsne}(d), a significant performance gain due to enhancement of separability for context $\Ca$ can be obtained without harming the model's performance (and separability of features among different classes) for the unseen context $\Cu$ as shown in \ref{fig:tsne}(h).

Thus, the level of feature separability across $\Ca$ and $\Cu$ for different model states reflects the generalizability improvements achieved through the steps of Algorithm \ref{alg:main}.

\section{Details of ablation study}
\label{ablation-study}
This section presents evaluations showing the effectiveness of the design choices of \textcolor{violet}{CRoP}, focusing on the WIDAR dataset in Scenario 1. Figure \ref{fig:ablation} compares the current design choices with alternative options available in the literature. The comparison is done for each of the three users chosen for personalization under both available $\Ca$ and unseen $\Cu$ contexts. The metric used for this comparison is the inference accuracy. Similar patterns were observed in other scenarios and datasets.

\subsection{Pruning mechanism}
\label{appendix:pruning-mechanism-ablation}

\textcolor{violet}{CRoP} uses one-shot magnitude-based pruning ($MP$) \citep{ThiNet-ICCV17,zhu2017prune} to remove the lowest-magnitude model parameters in the ToleratedPrune module (Algorithm \ref{alg:main}, Line 3). Various other pruning methods exist, most relevant ones being: Gradient-Based Pruning ($GP$) \citep{GP-soa}, pruning top-magnitude weights instead of lower ones ($MP-T$) \citep{bartoldson2020generalizationstability}, and iterative pruning ($MP-I$) \citep{paganini2020iterative} \citep{pruning-survey}. A comparative discussion of these methods is provided below.


\begin{wrapfigure}{r}{0.3\textwidth}
    \centering
    \subfloat[Pruning Mechanism]{\includegraphics[width=\linewidth]{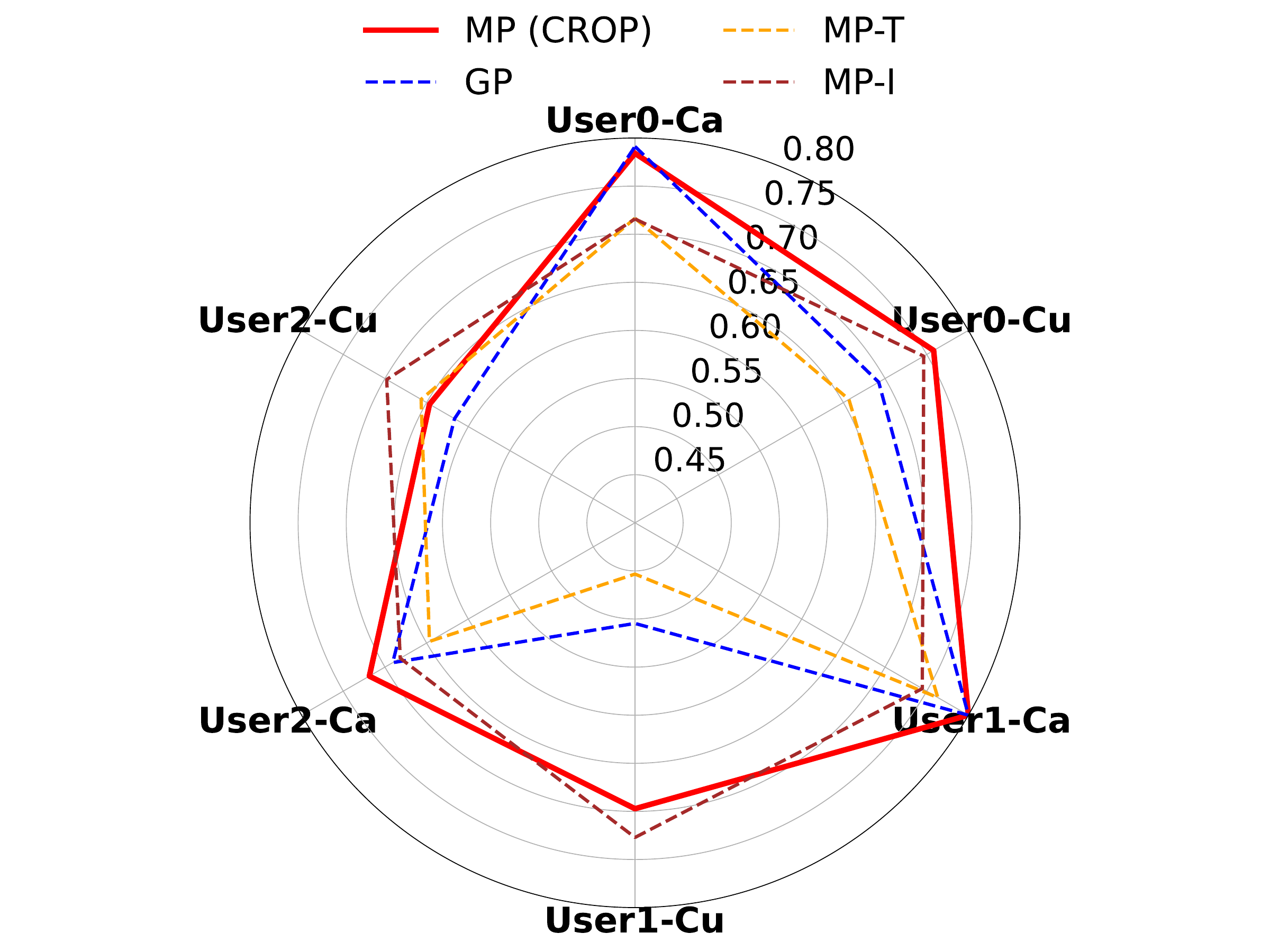}}
    \label{fig:pruningmechs}
    
    \subfloat[Regularization Mechanism]{\includegraphics[width=\linewidth]{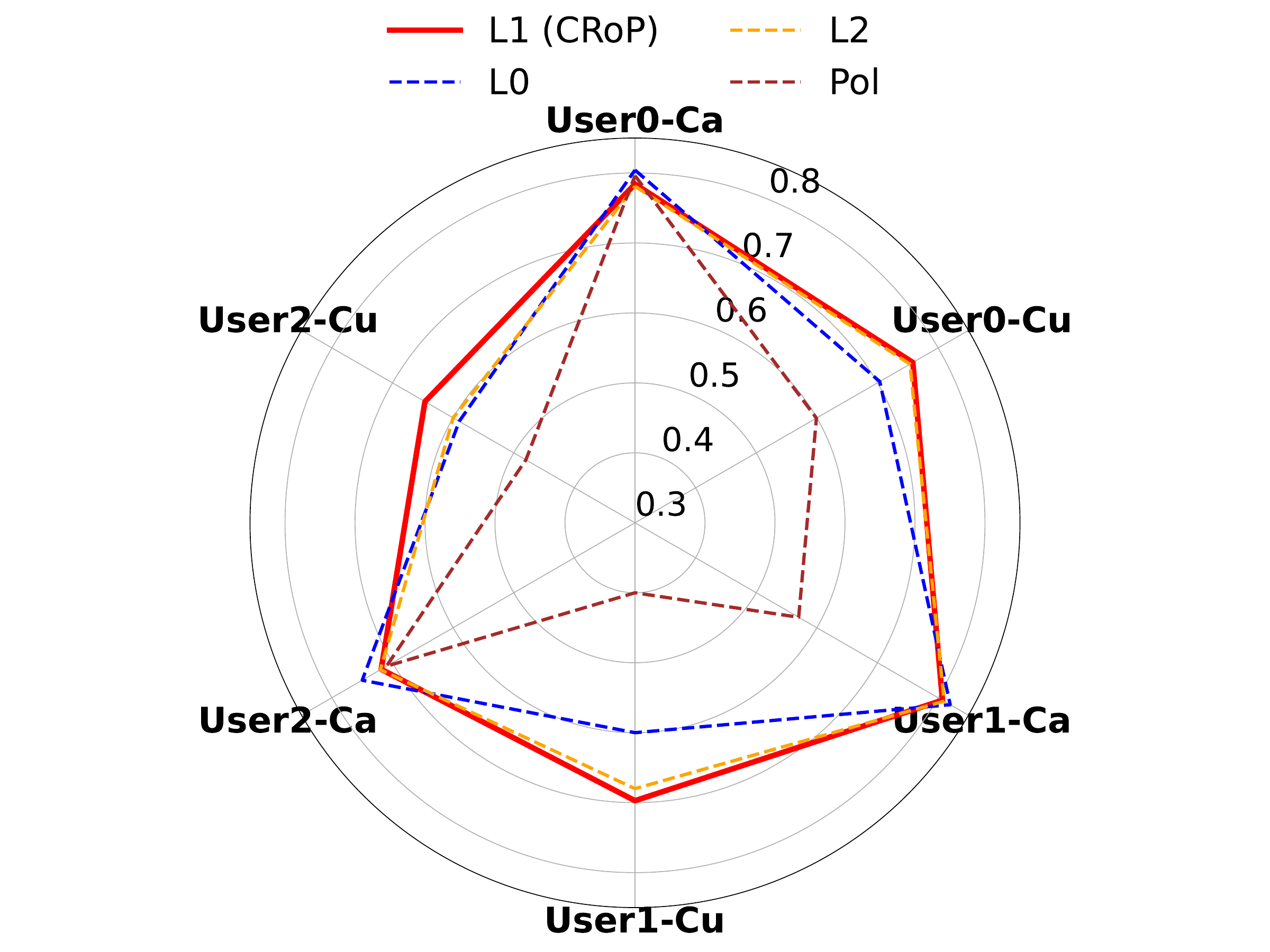}}
    \label{fig:regmechs}

    \caption{Ablation study exploring different alternatives for (a) pruning and (b) regularization mechanisms using accuracy as the metric of evaluation}
    \label{fig:ablation}
\vspace{-2ex}
\end{wrapfigure}

\textbf{Magnitude-based ($MP$) vs. Gradient-based Pruning ($GP$):} 
\citet{GP-soa} introduced gradient-based pruning ($GP$), which prunes parameters based on the $\ell_1$-norm of gradients. While effective for IID data (\citep{cifar10}), in \textcolor{violet}{CRoP}, where $\Ca$ and $\Cu$ have different distributions, $GP$ overfits to 
$\Ca$ and performs poorly on 
$\Cu$. In contrast, $MP$ selects parameters by weight magnitude, making it more robust and better performing on $\Cu$. As shown in Figures \ref{fig:ablation}(a), $GP$ performs well on $\Ca$ but consistently underperforms compared to $MP$ on $\Cu$ across all three users.

\textbf{Top prune ($MP-T$) vs. lower prune ($MP$):} 
\citet{bartoldson2020generalizationstability} suggests that pruning the top weights and retraining them ($MP-T$) can improve generalization by flattening loss landscapes. However, Figure \ref{fig:ablation}(a) shows that it performs significantly inferior to \textcolor{violet}{CRoP} for all users in both contexts.
This is because \textcolor{violet}{CRoP} aims to isolate a user-specific sub-network that performs well in the available context $\Ca$, while $MP-T$ removes important user-specific information. Moreover, the final finetuning stage (Algorithm \ref{alg:main} Line 5) retrains those pruned parameters highly important for $\Ca$, overwriting the generic information impacting context $\Cu$.

\textbf{One shot ($MP$) vs Iterative approach ($MP-I$)}
The presented approach, i.e., $MP$, uses a one-shot approach where the initially finetuned model undergoes one pass of pruning, mixing, and finetuning. However, we compared it with an iterative variation as well. To implement the iterative approach, $MP-I$, we allow the initially finetuned model to undergo multiple passes of pruning, mixing, and fine-tuning. As shown in Figure \ref{fig:ablation}(a), $MP-I$ does not significantly improve model performance, yet it incurs high computational costs through repetition.

In the one-shot approach, the `ToleratedPrune' module (Algorithm \ref{alg:main} line 3 and Algorithm \ref{alg:prune}) calculates the optimal pruning amount by removing the maximum parameters without exceeding the accuracy loss threshold $\tau$ for context $\Ca$. In the iterative approach, we gradually reach the same pruning amount by removing small fractions of parameters for each cycle. These parameters are replaced by weights from the generic model, but since the finetuning loss function resembles the initial one,  where no early stopping mechanism is used, it drives those weights to lower magnitudes, causing them to be pruned again in the next cycle. This repeats until the target pruning amount is reached, effectively pruning a similar set of weights as the one-shot method but in smaller steps.

\subsection{Regularization Mechanisms}
\label{appendix:regularization-ablation}
\textcolor{violet}{CRoP} uses regularization to push model parameters toward zero, making pruning easier in later steps. Weight penalty-based regularization methods apply different norms to the model's weights, such as $\ell_0$, $\ell_1$, $\ell_2$, and polarization. The $\ell_1$ norm is particularly effective because it not only reduces model parameters but also makes the least significant parameters to zero. This property also makes $\ell_1$ useful for feature selection \cite{8672565}.
Figure \ref{fig:ablation}(b) compares all these possible regularization choices. We observed that among $\ell_0$, $\ell_1$, $\ell_2$ and polarization \citep{polarization}, using $\ell_1$ regularization is most effective in the unseen context. This is because, by forcing the least important parameter weights to zero, it allows maximum pruning with `ToleratedPrune,' helping to recover as much generic information as possible.

\begin{table*}[!]
\resizebox{0.65\linewidth}{!}{
    \centering
    \begin{tabular}{|c|c|c|c|c|}
        \hline
         \textbf{Dataset} & \textbf{Platform} & \textbf{Train (s)} & \textbf{RSS (MBs)} & \textbf{ \makecell{ Resource \\Usage (\%)}}  \\
         \hline
          \multirow{7}{*}{PERCEPT-R} &AMD Ryzen 9 5950X 16-Core Processor&17.68&2881.13&23\\  \cline{2-5}
         &Apple M1 &119.56&288&1\\  \cline{2-5}
         &NVIDIA RTX A6000 &5.16&1108.65&16\\  
         \cline{2-5}
         &AMD Ryzen 9 7950X 16-Core Processor &16.82 &807.58&24\\  \cline{2-5}
         &NVIDIA GeForce RTX 4090&2.34&1166.82&23\\
         \cline{2-5}
         &Google Pixel 6&824.19&360.95&36.8\\
         \cline{2-5}
         &Cortex-A76 (Raspberry Pi 5) &190.44&505.67&49\\
         \hline
         \multirow{7}{*}{Stress-Sensing}&AMD Ryzen 9 5950X 16-Core Processor&8.77&1510.28&21\\  \cline{2-5}
         &Apple M1 &12.93&134&1\\ \cline{2-5}
         &NVIDIA RTX A6000 &11.34&1081.73&3\\  \cline{2-5}
         &AMD Ryzen 9 7950X 16-Core Processor & 3.68 &624.98 & 23\\  \cline{2-5}
         &NVIDIA GeForce RTX 4090& 5.01 & 1136.69 &16\\  
         \cline{2-5}
         &Google Pixel 6&98.45&246.06&33\\
         \cline{2-5}
         &Cortex-A76 (Raspberry Pi 5) &50.86&392.74&49\\
         \hline
    \end{tabular}}
    \caption{Resource requirement for one time training using \textcolor{violet}{CRoP}}
    \label{tab:resource-requirement}

\resizebox{0.9\linewidth}{!}{
    \centering
    \begin{tabular}{|c|c|c|c|c||c|c|c|}
        \hline
        &Model $\rightarrow$ & \multicolumn{3}{|c|}{Generic Model} & \multicolumn{3}{|c|}{Personalized Model} \\
        \hline
         \textbf{Dataset} & \textbf{Platform} & \textbf{Inference (s)} & \textbf{RSS (MBs)} & \textbf{ \makecell{ Resource \\Usage (\%)}}  & \textbf{Inference (s)} & \textbf{RSS (MBs)} & \textbf{ \makecell{ Resource \\Usage (\%)}}  \\
         \hline
          \multirow{7}{*}{PERCEPT-R} &AMD Ryzen 9 5950X 16-Core Processor&0.03&2752&16&0.03&2752&16\\  \cline{2-8}
         &Apple M1 &0.22&362.34&0.19&0.22&361&0.19\\  \cline{2-8}
         &NVIDIA RTX A6000 &0.09&773.33&3&0.09&778.25&1\\  
         \cline{2-8}
         &AMD Ryzen 9 7950X 16-Core Processor &0.04&697&12&0.02&697&12\\  \cline{2-8}
         &NVIDIA GeForce RTX 4090&0.07&829.5&0.01&0.07&829.75&0.01\\
         \cline{2-8}
         &Google Pixel 6&0.62&249.49&37.85&0.61&253.89&38.08\\
         \cline{2-8}
         &Cortex-A76 (Raspberry Pi 5) &0.2&392.01&49.84&0.18&392.57&49.8\\
         \hline
         \hline
         \multirow{7}{*}{Stress-Sensing}&AMD Ryzen 9 5950X 16-Core Processor&0.03&510&23&0.03&510&23\\  \cline{2-8}
         &Apple M1 &0.1&206&0.4&0.1&265&0.4\\ \cline{2-8}
         &NVIDIA RTX A6000 &0.1&752&2&0.1&753&1\\  \cline{2-8}
         &AMD Ryzen 9 7950X 16-Core Processor &0.01&537.14&22.7&0.01&537&22.7\\  \cline{2-8}
         &NVIDIA GeForce RTX 4090&0.01&800&1&0.01&800&1\\  
         \cline{2-8}
         &Google Pixel 6&0.13&201.24&25.62&0.11&201.4&29.03\\
         \cline{2-8}
         &Cortex-A76 (Raspberry Pi 5) &0.14&335.77&48.78&0.15&335.83&48.7\\
         \hline
    \end{tabular}}
    \caption{Runtime Resource requirement comparison for generic and personalized models obtained for \textcolor{violet}{CRoP}}
    \label{tab:resource-requirement-runtime}
    \vspace{-5ex}
\end{table*}

\subsection{Full finetune vs. partial finetune}
\label{appendix:full-vs-partial-ablation}
In order to keep the zeroed-out parameters as zero, conventional pruning approaches finetune only the weights that are retained during the pruning phase. However, in the presented approach, the zeroed-out weights are replaced by corresponding weights from the generic model. Thus, there are no zero parameters. So, the presented approach finetunes all the parameters. We still evaluated both finetuning approaches and observed that there is no significant difference between the two approaches. 

\section{Run-time analysis}
\label{runtime-analysis}
\textcolor{violet}{CRoP} is a static personalization approach that requires one-time on-device training of the generic model during personalization. We performed run-time evaluations to assess the viability of the \textcolor{violet}{CRoP} framework across \textit{seven} different deployment platforms: AMD Ryzen 9 5950X 16-Core Processor, AMD Ryzen 9 7950X 16-Core Processor, Apple M1, NVIDIA RTX A6000, NVIDIA GeForce RTX 4090,  and resource-constrained devices Google Pixel 6 and Cortex-A76 (Raspberry Pi 5). Table \ref{tab:resource-requirement} shows the average resources required in the training phase in terms of training time (s), process memory (MBs), and resource, i.e., GPU/CPU utilization (\%)  for two health-related datasets: PERCEPT-R and Stress-Sensing. Notably, the resource utilization reported in Tables \ref{tab:resource-requirement} and \ref{tab:resource-requirement-runtime} refers to GPU utilization for the NVIDIA RTX A6000 and NVIDIA GeForce RTX 4090 platforms, whereas for all other devices, it reflects CPU utilization.  According to our evaluation, except for the resource-constrained device, Apple M1, Google Pixel 6 and Raspberry Pi 5, the training time is a couple of seconds, and resource requirements are minimal. Even on resource-constrained devices like Apple M1, Google Pixel 6 and Raspberry Pi 5, computation time remains efficient, with only a slight increase, typically adding at most a few minutes. 
Since this is a one-time process, \textcolor{violet}{CRoP} is efficient compared to continuous learning methods that require repeated adaptation. Particularly, this makes \textcolor{violet}{CRoP} practical for critical applications, such as clinical settings, where it incurs only a brief, one-time computation overhead during enrollment.

Further, Table \ref{tab:resource-requirement-runtime} shows runtime comparison of the generic off-the-shelf model and the personalized model achieved using \textcolor{violet}{CRoP} accross seven devices mentioned above . It can be observed that inference time and resource usage for both model states are similar on all devices as the model architecture of the personalized model remains same as that of the off-the-shelf generic model. Thus, \textcolor{violet}{CRoP} does not introduce any inference time overheads.

\section{Discussion, Limitations and Future Direction}
\label{discussion}

\textbf{Analyzing the pruning pattern across all layers of the model.} We employ regularization (during the initial finetuning step) and pruning (during \emph{ToleratedPruned} step) to identify model parameters that are essential to represent user-specific traits present in the available context $\Ca$ data. The remaining parameter values are restored from the generic model to incorporate generic knowledge. 

Our analysis of the LeNet model trained on the WIDAR dataset revealed that the initial fine-tuning step has minimal impact on the initial feature extraction layers, as they primarily capture generic low-level features. In contrast, the deeper layers—particularly the last convolutional and first linear layers—undergo significant changes as they learn high-level feature representations essential for user-specific traits. Additionally, the regularization imposed by initial finetuning constrains these learned parameters to a small subset of the deeper layers, enabling the ToleratedPrune module to prune these layers more aggressively than the initial ones. Consequently, CRoP preserves low-level generic features by minimally affecting early layers during initial finetuning while allowing deeper layers to adapt to individual traits. Simultaneously, it integrates more generic information from the generic model during model mixing on those deeper layers by performing more aggressive pruning, ensuring a balance between personalization and generalization. The detailed discussion in provided in Appendix \ref{appendix:layer-wise-distance}.

\textbf{Evaluating different pruning paradigms} The paper performed a limited evaluation on pruning paradigms through ablation studies as it was not the primary focus of the study. Section \ref{ablation-study} on ablation study justifies \textcolor{violet}{CRoP}’s design choice but does not establish any particular paradigm’s superiority in unseen contexts.

\textbf{Dependence on quality of off-the-shelf generic model.} As the approach relies on using a pre-trained off-the-shelf model as an input,  the quality of this model can impact the performance of the final personalized models. As discussed in Section \ref{cross-context}, users with suboptimal performance of the generic model show higher gain through personalization. On the other hand, for certain participants in the PERCEPT-R dataset, a high-performing model leaves minimal room for improvement through personalization.

\textbf{Exploring different model architectures for each dataset.} We restrict our study to the models benchmarked and deployed for datasets used in this work without accounting for model variability. However, the different model architectures employed across all the datasets incorporate a range of layers such as convolutional, linear, BiLSTM, and GRU, which introduces some variability.

\textbf{Assessing the need for inter-user generalizability.} The inter-user heterogeneity \citep{gong2023dapper,meegahapola2023generalization, sempionatto2021wearable} kindles the need for personalization. To address this, several human sensing applications \citep{crosscheck,PTN-baseline,EMGSense,MobilePhys} adopt personalization to generate models tailored for user-specific traits. Notably, \citet{crosscheck} have shown that models trained for a set of users show higher mean average error for data belonging to new users due to inter-user heterogeneity.  This challenge parallels domain adaptation scenarios, where each individual can be viewed as a distinct domain, and models adapted to one domain do not transfer well to other domains \citep{adversarial-da}. Addressing the adaptation of a personalized model to a new user—i.e., inter-user generalizability—remains an open research question. In this work, we restrict our evaluations to the conventional definition of personalization and leave inter-user generalization for future exploration. Additionally, exploring the incremental addition of new users and comparing them against existing incremental learning methods is an interesting direction for future work. This would also allow evaluation of method’s ability to mitigate catastrophic forgetting, which is crucial for broader applicability.

\textbf{Trade-off between personalized model's performance in available $\Ca$ and unseen $\Cu$ context.} The goal of personalization in our work extends beyond improving performance in just the available context $\Ca$; it aims to enhance the overall performance of the model across all contexts ( that is, both available $\Ca$ and unseen $\Cu$ contexts). The conventional personalization approaches used as baselines in this work, often update the model using available context data $\Ca$ to the extent that results in significant performance drops in unseen contexts $\Cu$ and thus, limit the overall performance gains. In contrast, \textcolor{violet}{CRoP} achieves better balance by improving performance in the available context $\Ca$ while minimizing performance degradation in unseen contexts $\Cu$. This advantage is formalized in our problem statement and evaluation metric \emph{Personalization} ($\Delta_P$) and \emph{Generalization} ($\Delta_G$). Table \ref{tab:main-table} reports consistently positive $\Delta_P$ and $\Delta_G$ values achieved by \textcolor{violet}{CRoP} for all datasets under all scenarios, indicating that \textcolor{violet}{CRoP}-personalized models $\MP$ achieve overall performance improvements over the generic model $\MG$ and \emph{conventionally-finetuned} model $\MC$, respectively.

\section{Conclusion}
\label{sec:conclusion}
This study introduces \textcolor{violet}{CRoP}, a novel static personalization approach generating context-wise intra-user robust models from limited context data. Using pruning in a novel way to balance personalization and generalization, empirical analysis on four human-sensing datasets shows \textcolor{violet}{CRoP} models exhibit an average increase of $35.23\%$ in \emph{personalization} compared to generic models and $7.78\%$ in generalization compared to \emph{conventionally-finetuned} personalized models. \textcolor{violet}{CRoP} utilizes off-the-shelf models, reducing training effort and addressing privacy concerns. Since no steps rely on specific model components, \textcolor{violet}{CRoP} is architecture-agnostic, ensuring broad applicability. With practical benefits and quantitative performance enhancements, \textcolor{violet}{CRoP} facilitates reliable real-world deployment for AI-based human-sensing applications like healthcare.

\section*{Acknowledgements}
This work was partly supported by NSF IIS SCH \#2124285 and NIH R01 \#R01DC020959.



\bibliographystyle{plainnat}
\bibliography{bibfile}

\appendix

\section{Ensuring Reproducibility}
\label{reproducibility}

Depending on the distribution of the data, different accuracy measures have been used in the literature such as balanced accuracy, standard accuracy, or F1 score. To ensure consistency with the original baseline papers for each dataset \cite{Benway2023-eu, stress-sensing}, we follow their evaluation metric. Detailed explanations and justifications are provided in Appendix \ref{appendix:Metrics_for_classification_accuracy_evaluation}. To ensure reproducibility, we provide the hyperparameters used in both the general model training phase and the personalization phase in Appendix \ref{appendix:hyperparameters}. This includes learning rate, alpha, $\tau$, epoch count, and other settings specific to each dataset. Additionally Appendix \ref{appendix:code} and Appendix \ref{appendix:compute-resources} provide details of the code and the compute resources, respectively.

\subsection{Model Architectures}
\label{model-architectures}
For each dataset, we employ models used in the literature:

\paragraph{PERCEPT-R}: In line with the literature \citet{Benway2023-eu}, we tried several model architectures such as CNN, DNN, BILSTM, etc, whose number of parameters were identified using grid search. Among those, the biLSTM model containing 4 bidirectional LSTM layers followed by 5 linear layers, accompanied by a Hardswish activation layer, was identified as the one that exhibited the best results for the generic data and was used for this study.

\paragraph{WIDAR:} 
The model used for WIDAR follows the LeNet architecture \citep{WIDAR-dataset}, which contains three 2D convolutional layers followed by two linear layers. Each of these layers, except the final classification layer, is followed by a ReLU activation layer.

\paragraph{ExtraSensory:} 
The model follows a CNN-GRU-based architecture used in HAR literature \citep{metasense,invariant_feature_learning}. The model consists of three batch-normalized 1D convolution layers followed by a linear layer that feeds into a batch-normalized recursive (GRU) layer and two linear layers to generate embeddings. For the classification head, two linear layers were used.

\paragraph{Stress Sensing Dataset:} The model uses a simple multi-layer-perceptron (MLP) architecture \cite{eren2022stress} consisting of 3 linear layers with hidden size of 128.

\subsection{Metrics for classification accuracy evaluation}
\label{appendix:Metrics_for_classification_accuracy_evaluation}
We use accuracy to measure the performance of a model. However, the computation of this metric differs for the four datasets. The details of the metrics used for all the datasets are as follows:

\begin{enumerate}
    \item PERCEPT-R: For this dataset, \citet{Benway2023-eu} utilized balanced accuracy for the binary classification task, and we employed the same metrics in our study.
    \item WIDAR: We use a 6-class classification for gesture recognition, and the distribution of the data among these classes is nearly balanced. Thus, standard classification accuracy has been used for WIDAR.
    \item ExtraSensory: The subset of the Extrasensory dataset used for this work aims for a binary classification for activity recognition. We observed that the data distribution was quite imbalanced among the two classes, and therefore, balanced classification accuracy was used for this dataset. Balanced accuracy is computed as the average of true positive rate and true negative rate.
    \item Stress Sensing Dataset: For this binary classification problem, F1 score has been used as a performance metric as suggested by the original authors \cite{stress-sensing}.
    
\end{enumerate}
For simplicity, we use the term `accuracy' to encompass all the metrics discussed above.

\subsection{Hyperparameters}
\label{appendix:hyperparameters}

The approach uses several hyperparameters for generic model training and personalization. Table \ref{tab:hyperparams-global} and \ref{tab:hyperparams-personal} show the hyperparameter values for generic and personalized model training, respectively. These values correspond to the best results obtained for the data belonging to the available context using a grid search. For training the generic model, in addition to the number of epochs, `Base Learning Rate' and `Max Learning Rate' (the arguments for CycleLR \citep{cycleLR}) are the hyperparameters. For the personalized model, learning rate (fixed), $\alpha$, $\tau$, number of epochs for initial finetuning (Initial Epochs), and epochs for final finetuning (Final Epochs) are the hyperparameters. The range of these hyperparameters used for grid search during personalization is also mentioned in Table \ref{tab:hyperparams-personal}. 

\begin{table}[]
    \centering
    \resizebox{0.60\linewidth}{!}{
    \begin{tabular}{|c|c|c|c|c|}
        \hline
        Hyperparameter &PERCEPT-R& WIDAR & ExtraSensory & Stress-sensing \\
        \hline
         Base Learning Rate&1e-5&1e-07&1.2e-08&5e-5\\
         Max Learning Rate&1e-5&5e-06&7.5e-07&5e-5\\
         Epochs&300&1000&150&1000\\
         \hline
    \end{tabular}}
    \caption{Hyperparameters for generic Models}
    \label{tab:hyperparams-global}
\end{table}

\begin{table*}[]
    \centering
    \resizebox{0.8\linewidth}{!}{
    \begin{tabular}{|c|c|c|c|c|c|}
        \hline
        
        Hyperparameter &Range&PERCEPT-R& WIDAR & ExtraSensory& Stress Sensing\\
        \hline
         Learning Rate&1e-6 - 1e-1&1e-5&1e-6&1e-6& 1e-5\\
         $alpha$&1e-6 - 10&0.01&0.0001&0.5 & 0.0001\\
         $\tau$& 0.01 - 0.25 &0.05&0.2&0.01& 0.01\\
         Initial Epochs& 100 -1000&300&600&600 & 1000 \\
         Final Epochs&100 - 1000&300&600&1000& 1000 \\
         \hline
    \end{tabular}}
    \caption{Hyperparameters for Personalized Models}
    \label{tab:hyperparams-personal}
\end{table*}

Additionally, we use $k=k'=0.05$ for the \emph{ToleratedPrune} module for PERCEPT-R, WIDAR, and ExtraSensory datasets, while for the Stress-sensing dataset, $k = 0.05$ and $k'=0.01$ is being used. One may find different values to be suitable for other datasets and model architectures.

\subsection{Code}
\label{appendix:code}
The code is provided at this \href{https://github.com/Sawinder-Kaur/CRoP_IMWUT}{Git Repository} arranged into dataset-specific folders. Each folder contains the pre-trained generic model, all the required modules, and the instructions to run the code. The seed values used for the evaluations are also provided in the shell files. The data partitioned into personalized and context-wise sets will be released upon publication.

\subsection{Compute Resources}
\label{appendix:compute-resources}
All the computations have been performed on NVIDIA Quadro RTX 5000 with 48 RT Cores and 16GB
GDDR6 memory.

\section{Detailed user-specific results}
\label{appendix:detailed-results}
This section discusses the user-specific patterns. Appendix \ref{delta-p-details} discusses detailed personalization ($\Delta_P$) results while Appendix \ref{perosnalized-comparizon} discusses generalization ($\Delta_G$) results. Further, Appendix \ref{apendix:error-bars} shows person-wise standard deviation values for generic $\MG$, conventionally finetuned $\MC$ and \textcolor{violet}{CRoP} $\MP$ models.
\label{Appendix:detailed-results}

\subsection{Detailed discussion of $\Delta_P$ results}
\label{delta-p-details}
\par The personalized models obtained using \textcolor{violet}{CRoP} exhibit higher classification accuracy than the generic models on the available context's data $\Dai$, showcasing the benefits of personalization.
To demonstrate the existence of such improvement, Tables \ref{tab:Global-Context1-WIDAR}- \ref{tab:Global-Context2-EDA-2} and Table \ref{tab:global-percept-r} compare the performance of generic model $\MG$ and personalized models obtained using \textcolor{violet}{CRoP} $\MP$.

\paragraph{WIDAR:} Tables \ref{tab:Global-Context1-WIDAR} and \ref{tab:Global-Context2-WIDAR} show that there is an average improvement of $25.25$ and $11.88$ percent points among three users for the available context $\Ca$ for Scenario 1 and Scenario 2, respectively. However, this benefit comes at the cost of a reduction in accuracy for the unseen context. There is an average reduction of $16.69$ and $5.97$ percent points for Scenario 1 and Scenario 2, respectively, for the unseen context $\Cu$. Notably, the loss of accuracy in the unseen context is much lower as compared to the \emph{conventionally-finetune} model as discussed in the Section \ref{motivation} (Motivation).

\paragraph{ExtraSensory:} Similar patterns could be observed for the Extrasensory dataset. Tables \ref{tab:Global-Context1-extrasensory} and \ref{tab:Global-Context2-extrasensory} show that there is an average increment of $16.40$ and $18.37$ percent points for the available context for Scenario 1 and Scenario 2, respectively. Interestingly, the performance of the personalized model for Scenario 1 on unseen context $\Cu$ was not adversely impacted. This is attributed to the fact that the inertial sensing patterns of Bag and Pocket phone carrying modes capture the user's body movement, whereas the phone-in-hand movement patterns can be distinct. In Scenario 1, $\Ca$ comprises pocket and $\Cu$ comprises bag,  meaning both available and unseen contexts encompass similar inertial patterns, leading to advantageous performance even in the unseen context. \emph{This evaluation illustrates minimal intra-user generalizability loss on unseen contexts when both available and unseen contexts share similar user traits.} However, in Scenario 2, where only the hand belongs to the unseen context $\Cu$, there is an average loss of $5.02$ percentage points on the unseen context.


\paragraph{Stress Sensing:} The physiological features used in this dataset vary significantly from one user to other. Thus, Tables \ref{tab:Global-Context1-EDA-1}-\ref{tab:Global-Context2-EDA-2} show that the generic models do not perform well on personalized data. Personalized finetuning enables the model to learn user-specific patterns, allowing the model's performance to improve not only in the available context but also in the unseen context. This results in average personalization benefit $(\Delta_P)$ of $67.81$ and $85.25$ for Scenario 1 and Scenario 2, respectively. It is important to note that for each Scenario, only one model is trained for the available context and tested for two different unseen contexts. Moreover, double context change (Tables \ref{tab:Global-Context1-EDA-2} and \ref{tab:Global-Context2-EDA-2}) shows lower personalization benefit as compared to single context change (Tables \ref{tab:Global-Context1-EDA-1} and \ref{tab:Global-Context2-EDA-1}).

\paragraph{PERCEPT-R:} In this dataset, the heterogeneity of features among individuals is reflected through the difference in prediction accuracy of the generic model. It can be observed in Table \ref{tab:global-percept-r} that for some individuals, the generic model exhibits over $90\%$ accuracy on the available context data, while for others, the generic model's accuracy drops to around $60\%$. This results in significant variability over gains in available and unseen contexts. Overall, \textcolor{violet}{CRoP} yields an average personalization gain of $5.09\%$.

On average over all the datasets, a personalization benefit $(\Delta_P)$ of $35.23$ percent points are seen as compared to the generic models across the four datasets under both scenarios.

These evaluations establish that the personalized models obtained using \textcolor{violet}{CRoP} demonstrate improved performance over the available context data than the generic models and exhibit personalization.

\subsection{Detailed discussion of $\Delta_G$ results}
\label{perosnalized-comparizon}
\par The personalized models obtained using 
\textcolor{violet}{CRoP} ($\MP$) are expected to have higher accuracy on unseen context $\Cu$ than the \emph{conventionally-finetune} personalized models ($\MC$) as discussed in Section \ref{motivation}. This section assesses whether the results align with these expectations.

\paragraph{WIDAR:} Tables \ref{tab:Personal-Context1-WIDAR} and \ref{tab:Personal-Context2-WIDAR} demonstrate that the personalized models $\MP$ exhibit an average increment of $8.01$ and $2.85$ percent points in the unseen context for Scenario 1 and Scenario 2, respectively. However, an average loss of $1.57$ and an average gain of $1.44$ percent points in $\Ca's$ accuracy could be observed for Scenario 1 and Scenario 2, respectively.

\paragraph{Extrasensory:} Similar patterns could be observed for the ExtraSensory dataset where the average accuracy on the unseen context improved by $4.97$ and $12.61$ percentage points for Scenario 1 and Scenario 2 as shown in Tables \ref{tab:Global-Context1-extrasensory} and \ref{tab:Personal-Context2-extrasensory}, respectively. As expected, there is a loss of $5.43$ and $4.44$ percent points in the available contexts for Scenario 1 and Scenario 2, respectively. 


\paragraph{Stress Sensing:} As observed in Tables \ref{tab:Global-Context1-EDA-1}-\ref{tab:Global-Context2-EDA-2}, personalized finetuning improves models performance on unseen context as well, we can claim that there is some user-specific traits which are common in available and unseen context. While comparing our final models with \emph{conventionally-finetuned} models (Tables \ref{tab:Personal-Context1-EDA-1}-\ref{tab:Personal-Context2-EDA-2}), performance boost in both available and unseen context could be observed. This can be attributed to the generalization improvement benefits of model pruning \citep{jin2022prunings}. This results in average generalization benefit $(\Delta_G)$ of $13.81$ and $13.08$ for Scenario 1 and Scenario 2, respectively, for single context change. Similar personalization benefits could be seen for double context change.


\paragraph{PERCEPT-R:} As observed in Table \ref{tab:personal-percept-r}, the variability in generalization benefits among different individuals is less pronounced as compared to personalization benefits. On average, \textcolor{violet}{CRoP} introduces a generalization benefit of $2.57\%$. 

On average over all the datasets, a generalization benefit $(\Delta_G)$ of $7.78\%$ percent points are seen over the \emph{conventionally-finetuned} personalized models across all datasets under both scenarios.



\begin{table*}[t]
\subfloat[Scenario 1 for WIDAR dataset]{
\resizebox{0.45\linewidth}{!}{
    \begin{tabular}{|c|cc|cc|cc|}
        \hline
        Model& \multicolumn{2}{|c|}{$\MG$} & \multicolumn{2}{|c|}{$\MP$} &\multicolumn{2}{|c|}{$\cA(\MP,\cC) - \cA(\MG,\cC)$}\\
        \hline
         User& $\Ca$& $\Cu$& $\Ca$& $\Cu$& $\Ca$& $\Cu$ \\
         \hline
         0&63.90&77.09&83.67&69.53&\textcolor{cyan}{+19.77}&\textcolor{orange}{-7.56}\\
         1&61.80&79.78&86.41&54.45&\textcolor{cyan}{+24.61}&\textcolor{orange}{-25.33}\\
         2&45.63&79.81&77.02&62.63&\textcolor{cyan}{+31.38}&\textcolor{orange}{-17.18}\\
         \hline
         Average&&&&&\textcolor{cyan}{+25.25}& \textcolor{orange}{-16.69}\\
         \hline
         $\Delta_P$&&&&&\multicolumn{2}{|c|}{\textcolor{cyan}{+8.55}}\\
         \hline
    \end{tabular}}
    \label{tab:Global-Context1-WIDAR}}
    \hfill
    \subfloat[Scenario 2 for WIDAR dataset]{
    \resizebox{0.45\linewidth}{!}{
    \begin{tabular}{|c|cc|cc|cc|}
        \hline
        Model& \multicolumn{2}{|c|}{$\MG$} & \multicolumn{2}{|c|}{$\MP$} &\multicolumn{2}{|c|}{$\cA(\MP,\cC) - \cA(\MG,\cC)$}\\
        \hline
         User& $\Ca$& $\Cu$& $\Ca$& $\Cu$& $\Ca$& $\Cu$ \\
         \hline
         0&73.28&61.80&82.59&58.38&\textcolor{cyan}{+9.31}&\textcolor{orange}{-2.43}\\
         1&73.18&59.58&92.44&47.90&\textcolor{cyan}{+19.27}&\textcolor{orange}{-11.67}\\
         2&80.45&46.13&87.5&42.31&\textcolor{cyan}{+7.04}&\textcolor{orange}{-3.81}\\
         \hline
         Average&&&&&\textcolor{cyan}{+11.88}&\textcolor{orange}{-5.97}\\
         \hline
         $\Delta_P$&&&&&\multicolumn{2}{|c|}{\textcolor{cyan}{+5.90}}\\
         \hline
    \end{tabular}}
    \label{tab:Global-Context2-WIDAR}}

\vspace{6pt}
\subfloat[Scenario 1 for ExtraSensory dataset]{
\resizebox{0.45\linewidth}{!}{
    \begin{tabular}{|c|cc|cc|cc|}
        \hline
        Model& \multicolumn{2}{|c|}{$\MG$} & \multicolumn{2}{|c|}{$\MP$} &\multicolumn{2}{|c|}{$\cA(\MP,\cC) - \cA(\MG,\cC)$}\\
        \hline
         User& $\Ca$& $\Cu$& $\Ca$& $\Cu$& $\Ca$& $\Cu$ \\
         \hline
         61&78.69&69.83&82.59&69.66&\textcolor{cyan}{+3.9}&\textcolor{cyan}{-0.17}\\
         7C&78.91&76.41&88.00&71.63&\textcolor{cyan}{+9.09}&\textcolor{orange}{-4.78}\\
         80&55.84&26.24&82.36&38.87&\textcolor{cyan}{+26.52}&\textcolor{cyan}{+12.63}\\
         9D&73.74&85.63&82.81&84.72&\textcolor{cyan}{+9.07}&\textcolor{orange}{-0.91}\\
         B7&56.06&88.33&89.50&86.97&\textcolor{cyan}{+33.44}&\textcolor{orange}{-1.36}\\
         \hline
         Average&&&&&\textcolor{cyan}{+16.40}&\textcolor{cyan}{+1.08}\\
         \hline
         $\Delta_P$&&&&&\multicolumn{2}{|c|}{\textcolor{cyan}{+17.49}}\\
         \hline
    \end{tabular}}
    \label{tab:Global-Context1-extrasensory}}
    \hfill
    \subfloat[Scenario 2 for ExtraSensory dataset]{
    \resizebox{0.45\linewidth}{!}{
    \begin{tabular}{|c|cc|cc|cc|}
        \hline
        Model& \multicolumn{2}{|c|}{$\MG$} & \multicolumn{2}{|c|}{$\MP$} &\multicolumn{2}{|c|}{$\cA(\MP,\cC) - \cA(\MG,\cC)$}\\
        \hline
         User& $\Ca$& $\Cu$& $\Ca$& $\Cu$& $\Ca$& $\Cu$ \\
         \hline
         61&76.43&80.00&87.24&73.44&\textcolor{cyan}{+10.81}&\textcolor{orange}{-6.56}\\
         7C&75.07&92.32&83.18&89.39&\textcolor{cyan}{+8.11}&\textcolor{orange}{-2.93}\\
         80&54.40&88.77&84.49&81.12&\textcolor{cyan}{+30.09}&\textcolor{orange}{-7.65}\\
         9D&75.58&75.02&82.58&74.65&\textcolor{cyan}{+7.00}&\textcolor{orange}{-0.37}\\
         B7&59.73&84.58&95.56&77.01&\textcolor{cyan}{+35.83}&\textcolor{orange}{-7.57}\\
         \hline
         Average&&&&&\textcolor{cyan}{+18.37}&\textcolor{orange}{-5.02}\\
         \hline
         $\Delta_P$&&&&&\multicolumn{2}{|c|}{\textcolor{cyan}{+13.35}}\\
         \hline
    \end{tabular}}
    \label{tab:Global-Context2-extrasensory}}

\vspace{6pt}
\subfloat[Scenario 1 for Stress Sensing - single context change]{
\resizebox{0.45\linewidth}{!}{
    \begin{tabular}{|c|cc|cc|cc|}
        \hline
        Model& \multicolumn{2}{|c|}{$\MG$} & \multicolumn{2}{|c|}{$\MP$} &\multicolumn{2}{|c|}{$\cA(\MP,\cC) - \cA(\MG,\cC)$}\\
        \hline
         User& $\Ca$& $\Cu$& $\Ca$& $\Cu$& $\Ca$& $\Cu$ \\
         \hline
         
         1&88.39&81.90&94.54&97.59&\textcolor{cyan}{+6.15}&\textcolor{cyan}{+15.69}\\
         2&47.40&50.0&77.12&90.47&\textcolor{cyan}{+29.72}&\textcolor{cyan}{+40.47}\\
         3&36.90&43.48&96.36&95.31&\textcolor{cyan}{+59.46}&\textcolor{cyan}{+51.93}\\
         \hline
         Average&&&&&\textcolor{cyan}{+31.78}& \textcolor{cyan}{+36.03}\\
         \hline
         $\Delta_P$&&&&&\multicolumn{2}{|c|}{\textcolor{cyan}{+67.81}}\\
         \hline
    \end{tabular}}
    \label{tab:Global-Context1-EDA-1}}
    \hfill
    \subfloat[Scenario 2 for Stress Sensing - single context change]{
    \resizebox{0.45\linewidth}{!}{
    \begin{tabular}{|c|cc|cc|cc|}
        \hline
        Model& \multicolumn{2}{|c|}{$\MG$} & \multicolumn{2}{|c|}{$\MP$} &\multicolumn{2}{|c|}{$\cA(\MP,\cC) - \cA(\MG,\cC)$}\\
        \hline
         User& $\Ca$& $\Cu$& $\Ca$& $\Cu$& $\Ca$& $\Cu$ \\
         \hline
         1&66.54&64.71&92.38&94.54&\textcolor{cyan}{+25.84}&\textcolor{cyan}{+29.83}\\
         2&69.10&50.70&85.26&89.65&\textcolor{cyan}{+16.16}&\textcolor{cyan}{+38.95}\\
         3&4.76&11.94&74.40&87.28&\textcolor{cyan}{+69.64}&\textcolor{cyan}{+75.34}\\
         \hline
         Average&&&&&\textcolor{cyan}{+37.21}&\textcolor{cyan}{+48.04}\\
         \hline
         $\Delta_P$&&&&&\multicolumn{2}{|c|}{\textcolor{cyan}{+85.25}}\\
         \hline
    \end{tabular}}
    \label{tab:Global-Context2-EDA-1}}

\vspace{6pt}
\subfloat[Scenario 1 for Stress Sensing - double context change]{
\resizebox{0.45\linewidth}{!}{
    \begin{tabular}{|c|cc|cc|cc|}
        \hline
        Model& \multicolumn{2}{|c|}{$\MG$} & \multicolumn{2}{|c|}{$\MP$} &\multicolumn{2}{|c|}{$\cA(\MP,\cC) - \cA(\MG,\cC)$}\\
        \hline
         User& $\Ca$& $\Cu$& $\Ca$& $\Cu$& $\Ca$& $\Cu$ \\
         \hline
         
         1&88.39&64.71&94.54&76.46&\textcolor{cyan}{+6.15}&\textcolor{cyan}{+11.75}\\
         2&47.40&50.70&77.12&63.22&\textcolor{cyan}{+29.72}&\textcolor{cyan}{+12.52}\\
         3&36.90&11.94&96.36&55.48&\textcolor{cyan}{+59.46}&\textcolor{cyan}{+43.54}\\
         \hline
         Average&&&&&\textcolor{cyan}{+31.78}& \textcolor{cyan}{+22.60}\\
         \hline
         $\Delta_P$&&&&&\multicolumn{2}{|c|}{\textcolor{cyan}{+54.38}}\\
         \hline
    \end{tabular}}
    \label{tab:Global-Context1-EDA-2}}
    \hfill
    \subfloat[Scenario 2 for Stress Sensing - double context change]{
    \resizebox{0.45\linewidth}{!}{
    \begin{tabular}{|c|cc|cc|cc|}
        \hline
        Model& \multicolumn{2}{|c|}{$\MG$} & \multicolumn{2}{|c|}{$\MP$} &\multicolumn{2}{|c|}{$\cA(\MP,\cC) - \cA(\MG,\cC)$}\\
        \hline
         User& $\Ca$& $\Cu$& $\Ca$& $\Cu$& $\Ca$& $\Cu$ \\
         \hline
         1&66.54&81.90&92.38&91.07&\textcolor{cyan}{+25.84}&\textcolor{cyan}{+9.17}\\
         2&69.10&50.00&85.26&62.84&\textcolor{cyan}{+16.16}&\textcolor{cyan}{+12.84}\\
         3&4.76&43.47&74.40&87.45&\textcolor{cyan}{+69.64}&\textcolor{cyan}{+43.98}\\
         \hline
         Average&&&&&\textcolor{cyan}{+37.21}&\textcolor{cyan}{+22.00}\\
         \hline
         $\Delta_P$&&&&&\multicolumn{2}{|c|}{\textcolor{cyan}{+59.21}}\\
         \hline
    \end{tabular}}
    \label{tab:Global-Context2-EDA-2}}

    \caption{Detailed Personalization ($\Delta_P$) results for WIDAR, ExtraSensory and  Stress Sensing dataset }
    \label{tab:Personalization}
\end{table*}

\begin{table*}[t]
\subfloat[Scenario 1 for WIDAR dataset]{
\resizebox{0.45\linewidth}{!}{
    \begin{tabular}{|c|cc|cc|cc|}
        \hline
        Model& \multicolumn{2}{|c|}{$\MC$} & \multicolumn{2}{|c|}{$\MP$} &\multicolumn{2}{|c|}{$\cA(\MP,\cC) - \cA(\MC,\cC)$}\\
        \hline
         User& $\Ca$& $\Cu$& $\Ca$& $\Cu$& $\Ca$& $\Cu$ \\
         \hline
         0&87.06&65.02&83.67&69.53&\textcolor{orange}{-3.38}&\textcolor{cyan}{+4.5}\\
         1&89.38&44.38&86.41&54.45&\textcolor{orange}{-2.97}&\textcolor{cyan}{+10.07}\\
         2&75.39&53.19&77.02&62.63&\textcolor{cyan}{+1.63}&\textcolor{cyan}{+9.44}\\
         \hline
         Average&&&&&\textcolor{orange}{-1.57}&\textcolor{cyan}{+8.01}\\
         \hline
         $\Delta_G$&&&&&\multicolumn{2}{|c|}{\textcolor{cyan}{+6.43}}\\
         \hline
    \end{tabular}}
    \label{tab:Personal-Context1-WIDAR}}
    \hfill
    \subfloat[Scenario 2 for WIDAR dataset]{
    \resizebox{0.45\linewidth}{!}{
    \begin{tabular}{|c|cc|cc|cc|}
        \hline
        Model& \multicolumn{2}{|c|}{$\MC$} & \multicolumn{2}{|c|}{$\MP$} &\multicolumn{2}{|c|}{$\cA(\MP,\cC) - \cA(\MC,\cC)$}\\
        \hline
         User& $\Ca$& $\Cu$& $\Ca$& $\Cu$& $\Ca$& $\Cu$ \\
         \hline
         0&77.30&57.46&82.59&58.38&\textcolor{cyan}{+5.29}&\textcolor{cyan}{+0.92}\\
         1&93.75&42.38&92.45&47.90&\textcolor{orange}{-1.30}&\textcolor{cyan}{+5.51}\\
         2&87.15&40.19&87.5&42.31&\textcolor{cyan}{+0.34}&\textcolor{cyan}{+2.13}\\
         \hline
         Average&&&&&\textcolor{cyan}{+1.44}&\textcolor{cyan}{+2.85}\\
         \hline
         $\Delta_G$&&&&&\multicolumn{2}{|c|}{\textcolor{cyan}{+4.30}}\\
         \hline
    \end{tabular}}
    \label{tab:Personal-Context2-WIDAR}}

\vspace{6pt}
\subfloat[Scenario 1 for ExtraSensory dataset]{
\resizebox{0.45\linewidth}{!}{
    \begin{tabular}{|c|cc|cc|cc|}
        \hline
        Model& \multicolumn{2}{|c|}{$\MC$} & \multicolumn{2}{|c|}{$\MP$} &\multicolumn{2}{|c|}{$\cA(\MP,\cC) - \cA(\MC,\cC)$}\\
        \hline
         User& $\Ca$& $\Cu$& $\Ca$& $\Cu$& $\Ca$& $\Cu$ \\
         \hline
         61&88.99&68.09&82.59&69.66&\textcolor{orange}{-6.40}&\textcolor{cyan}{+1.57}\\
         7C&92.58&61.74&88.0&71.63&\textcolor{orange}{-4.58}&\textcolor{cyan}{+9.89}\\
         80&86.51&49.82&82.36&38.87&\textcolor{orange}{-4.14}&\textcolor{orange}{-10.95}\\
         9D&88.89&83.14&82.81&84.73&\textcolor{orange}{-6.07}&\textcolor{cyan}{+1.58}\\
         B7&95.44&64.19&89.50&86.97&\textcolor{orange}{-5.94}&\textcolor{cyan}{+22.78}\\
         \hline
         Average&&&&&\textcolor{orange}{-5.43}&\textcolor{cyan}{+4.97}\\
         \hline
         $\Delta_G$&&&&&\multicolumn{2}{|c|}{\textcolor{orange}{-0.46}}\\
         \hline
    \end{tabular}}
    \label{tab:Personal-Context1-extrasensory}}
    \hfill
    \subfloat[Scenario 2 for ExtraSensory dataset]{
    \resizebox{0.45\linewidth}{!}{
    \begin{tabular}{|c|cc|cc|cc|}
        \hline
        Model& \multicolumn{2}{|c|}{$\MC$} & \multicolumn{2}{|c|}{$\MP$} &\multicolumn{2}{|c|}{$\cA(\MP,\cC) - \cA(\MC,\cC)$}\\
        \hline
         User& $\Ca$& $\Cu$& $\Ca$& $\Cu$& $\Ca$& $\Cu$ \\
         \hline
         61&93.90&64.27&87.24&73.47&\textcolor{orange}{-6.66}&\textcolor{cyan}{+9.17}\\
         7C&89.19&57.13&83.13&89.39&\textcolor{orange}{-6.01}&\textcolor{cyan}{+32.26}\\
         80&89.34&70.23&84.49&81.12&\textcolor{orange}{-4.85}&\textcolor{cyan}{+10.89}\\
         9D&85.53&72.95&82.58&74.65&\textcolor{orange}{-2.95}&\textcolor{cyan}{+1.7}\\
         B7&97.30&67.99&95.56&77.01&\textcolor{orange}{-1.74}&\textcolor{cyan}{+9.02}\\
         \hline
         Average&&&&&\textcolor{orange}{-4.44}&\textcolor{cyan}{+12.61}\\
         \hline
         $\Delta_G$&&&&&\multicolumn{2}{|c|}{\textcolor{cyan}{+8.17}}\\
         \hline
    \end{tabular}}
    \label{tab:Personal-Context2-extrasensory}}

\vspace{6pt}
\subfloat[Scenario 1 for Stress Sensing - single context change]{
\resizebox{0.45\linewidth}{!}{
    \begin{tabular}{|c|cc|cc|cc|}
        \hline
        Model& \multicolumn{2}{|c|}{$\MC$} & \multicolumn{2}{|c|}{$\MP$} &\multicolumn{2}{|c|}{$\cA(\MP,\cC) - \cA(\MC,\cC)$}\\
        \hline
         User& $\Ca$& $\Cu$& $\Ca$& $\Cu$& $\Ca$& $\Cu$ \\
         \hline
         
         1&91.17&92.15&94.54&97.59&\textcolor{cyan}{+3.37}&\textcolor{cyan}{+5.44}\\
         2&68.93&82.81&77.12&90.48&\textcolor{cyan}{+8.19}&\textcolor{cyan}{+7.67}\\
         3&84.78&90.13&96.36&95.31&\textcolor{cyan}{+11.58}&\textcolor{cyan}{+5.18 }\\
         \hline
         Average&&&&&\textcolor{cyan}{+7.71}& \textcolor{cyan}{+6.10}\\
         \hline
         $\Delta_G$&&&&&\multicolumn{2}{|c|}{\textcolor{cyan}{+13.81}}\\
         \hline
    \end{tabular}}
    \label{tab:Personal-Context1-EDA-1}}
    \hfill
    \subfloat[Scenario 2 for Stress Sensing - single context change]{
    \resizebox{0.45\linewidth}{!}{
    \begin{tabular}{|c|cc|cc|cc|}
        \hline
        Model& \multicolumn{2}{|c|}{$\MC$} & \multicolumn{2}{|c|}{$\MP$} &\multicolumn{2}{|c|}{$\cA(\MP,\cC) - \cA(\MC,\cC)$}\\
        \hline
         User& $\Ca$& $\Cu$& $\Ca$& $\Cu$& $\Ca$& $\Cu$ \\
         \hline
         1&92.37&96.46&92.38&94.54&\textcolor{cyan}{+0.01}&\textcolor{orange}{-1.93}\\
         2&75.57&72.47&85.26&89.65&\textcolor{cyan}{+9.69}&\textcolor{cyan}{+17.18}\\
         3&64.86&82.52&74.40&87.28&\textcolor{cyan}{+9.54}&\textcolor{cyan}{+4.76}\\
         \hline
         Average&&&&&\textcolor{cyan}{+6.41}&\textcolor{cyan}{+6.67}\\
         \hline
         $\Delta_G$&&&&&\multicolumn{2}{|c|}{\textcolor{cyan}{+13.08}}\\
         \hline
    \end{tabular}}
    \label{tab:Personal-Context2-EDA-1}}

\vspace{6pt}
\subfloat[Scenario 1 for Stress Sensing - double context change]{
\resizebox{0.45\linewidth}{!}{
    \begin{tabular}{|c|cc|cc|cc|}
        \hline
        Model& \multicolumn{2}{|c|}{$\MC$} & \multicolumn{2}{|c|}{$\MP$} &\multicolumn{2}{|c|}{$\cA(\MP,\cC) - \cA(\MC,\cC)$}\\
        \hline
         User& $\Ca$& $\Cu$& $\Ca$& $\Cu$& $\Ca$& $\Cu$ \\
         \hline
         
         1&91.17&72.10&94.544&76.46&\textcolor{cyan}{+3.37}&\textcolor{cyan}{+4.36}\\
         2&68.93&64.13&77.12&63.22&\textcolor{cyan}{+8.19}&\textcolor{orange}{-0.91}\\
         3&84.78&46.06&96.36&55.48&\textcolor{cyan}{+11.58}&\textcolor{cyan}{+9.42}\\
         \hline
         Average&&&&&\textcolor{cyan}{+7.71}& \textcolor{cyan}{+4.29}\\
         \hline
         $\Delta_G$&&&&&\multicolumn{2}{|c|}{\textcolor{cyan}{+12.00}}\\
         \hline
    \end{tabular}}
    \label{tab:Personal-Context1-EDA-2}}
    \hfill
    \subfloat[Scenario 2 for Stress Sensing - double context change]{
    \resizebox{0.45\linewidth}{!}{
    \begin{tabular}{|c|cc|cc|cc|}
        \hline
        Model& \multicolumn{2}{|c|}{$\MC$} & \multicolumn{2}{|c|}{$\MP$} &\multicolumn{2}{|c|}{$\cA(\MP,\cC) - \cA(\MC,\cC)$}\\
        \hline
         User& $\Ca$& $\Cu$& $\Ca$& $\Cu$& $\Ca$& $\Cu$ \\
         \hline
         1&92.37&87.22&92.38&91.07&\textcolor{cyan}{+0.01}&\textcolor{cyan}{+3.85}\\
         2&75.57&59.41&85.26&62.84&\textcolor{cyan}{+9.69}&\textcolor{cyan}{+3.43}\\
         3&64.86&83.49&74.40&87.45&\textcolor{cyan}{+9.54}&\textcolor{cyan}{+3.75}\\
         \hline
         Average&&&&&\textcolor{cyan}{+6.41}&\textcolor{cyan}{+3.75}\\
         \hline
         $\Delta_G$&&&&&\multicolumn{2}{|c|}{\textcolor{cyan}{+10.16}}\\
         \hline
    \end{tabular}}
    \label{tab:Personal-Context2-EDA-2}}
 \caption{Detailed Generalization ($\Delta_G$) results for WIDAR  ExtraSensory and Stress Sensing datasets }
 \label{tab:Generalization}
\end{table*}

\begin{table*}
    \subfloat[Personalization for PERCEPT-R]{
    \resizebox{0.45\linewidth}{!}{
    \begin{tabular}{|c|cc|cc|cc|}
         \hline
        Model& \multicolumn{2}{|c|}{$\MG$} & \multicolumn{2}{|c|}{$\MP$} &\multicolumn{2}{|c|}{$\cA(\MP,\cC) - \cA(\MG,\cC)$}\\
        \hline
         User& $\Ca$& $\Cu$& $\Ca$& $\Cu$& $\Ca$& $\Cu$ \\
         \hline
         17&72.72&60.51&73.30&63.19&\textcolor{cyan}{+0.58}&\textcolor{cyan}{+2.68}\\
         25&96.79&87.16&96.30&86.83&\textcolor{orange}{-0.49}&\textcolor{orange}{-0.33}\\
         28&55.22&54.88&67.12&60.17&\textcolor{cyan}{+11.90}&\textcolor{cyan}{+5.29}\\
         336&100&82.03&100&66.69&\textcolor{cyan}{0}&\textcolor{orange}{-15.34}\\
         344&77.54&67.48&81.9&65.36&\textcolor{cyan}{+4.36}&\textcolor{orange}{-2.12}\\
         361&63.63&89.93&64.28&81.73&\textcolor{cyan}{+0.64}&\textcolor{orange}{-8.2}\\
         362&59.03&66.23&78.52&82.51&\textcolor{cyan}{+19.49}&\textcolor{cyan}{+16.28}\\
         55&95.77&85.34&97.14&80.41&\textcolor{cyan}{+1.38}&\textcolor{orange}{-4.93}\\
         586&65.85&58.25&73.17&65.87&\textcolor{cyan}{+7.32}&\textcolor{cyan}{+7.62}\\
         587&64.71&65.1&69.4&65.19&\textcolor{cyan}{+4.68}&\textcolor{cyan}{+0.09}\\
         589&66.34&60.87&63.69&62.77&\textcolor{orange}{-2.64}&\textcolor{cyan}{+1.90}\\
         590&69.05&61.04&71.08&73.26&\textcolor{cyan}{+2.03}&\textcolor{cyan}{+12.22}\\
         591&61.91&58.68&72.03&63.44&\textcolor{cyan}{+10.12}&\textcolor{cyan}{+4.76}\\
         61&72.86&69.42&77.78&66.66&\textcolor{cyan}{+4.92}&\textcolor{orange}{-2.76}\\
         67&80.12&77.64&81.00&72.48&\textcolor{cyan}{+0.89}&\textcolor{orange}{-5.15}\\
         80&89.38&85.54&91.22&87.87&\textcolor{cyan}{+1.85}&\textcolor{cyan}{+2.33}\\
         \hline
         Average&&&&&\textcolor{cyan}{+4.19}&\textcolor{cyan}{+0.90}\\
         \hline
         $\Delta_P$&&&&&\multicolumn{2}{|c|}{\textcolor{cyan}{+5.09}}\\
         \hline
    \end{tabular}}
    \label{tab:global-percept-r}}
    \hfill
    \subfloat[Generalization for PERCEPT-R]{
    \resizebox{0.45\linewidth}{!}{
    \begin{tabular}{|c|cc|cc|cc|}
         \hline
        Model& \multicolumn{2}{|c|}{$\MC$} & \multicolumn{2}{|c|}{$\MP$} &\multicolumn{2}{|c|}{$\cA(\MP,\cC) - \cA(\MC,\cC)$}\\
        \hline
         User& $\Ca$& $\Cu$& $\Ca$& $\Cu$& $\Ca$& $\Cu$ \\
         \hline
         17&70.56&60.17&73.30&63.19&\textcolor{cyan}{+2.74}&\textcolor{cyan}{+3.01}\\
         25&94.91&83.8&96.30&86.83&\textcolor{cyan}{+1.39}&\textcolor{cyan}{+3.03}\\
         28&64.81&58.84&67.12&60.17&\textcolor{cyan}{+2.31}&\textcolor{cyan}{+1.33}\\
         336&100&60.79&100&66.69&\textcolor{cyan}{+0}&\textcolor{cyan}{+5.90}\\
         344&82.77&64.68&81.9&65.36&\textcolor{cyan}{-0.88}&\textcolor{cyan}{+0.68}\\
         361&57.61&78.66&64.28&81.73&\textcolor{cyan}{+6.67}&\textcolor{cyan}{+3.07}\\
         362&75.44&82.9&78.52&82.51&\textcolor{cyan}{+3.08}&\textcolor{orange}{-0.39}\\
         55&100&76.86&97.14&80.41&\textcolor{orange}{-2.86}&\textcolor{cyan}{+3.55}\\
         586&70.94&64.46&73.17&65.87&\textcolor{cyan}{+2.23}&\textcolor{cyan}{+1.41}\\
         587&70.84&66.23&69.4&65.19&\textcolor{orange}{-1.44}&\textcolor{orange}{-1.04}\\
         589&67.26&60.59&63.69&62.77&\textcolor{orange}{-3.57}&\textcolor{cyan}{+2.18}\\
         590&67.76&70.54&71.08&73.26&\textcolor{cyan}{+3.32}&\textcolor{cyan}{+2.73}\\
         591&71.83&64.2&72.03&63.44&\textcolor{cyan}{+0.20}&\textcolor{orange}{-0.76}\\
         61&74.93&65.23&77.78&66.66&\textcolor{cyan}{+2.86}&\textcolor{cyan}{+1.43}\\
         67&79.83&74.16&81.00&72.48&\textcolor{cyan}{+1.18}&\textcolor{orange}{-1.68}\\
         80&89.31&90.33&91.22&87.87&\textcolor{cyan}{+1.92}&\textcolor{orange}{-1.68}\\
         \hline
         Average&&&&&\textcolor{cyan}{+1.20}&\textcolor{cyan}{+1.37}\\
         \hline
         $\Delta_G$&&&&&\multicolumn{2}{|c|}{\textcolor{cyan}{+2.57}}\\
         \hline
    \end{tabular}}
    \label{tab:personal-percept-r}}
    \caption{Detailed Personalization ($\Delta_P$)and Generalization ($\Delta_G$)  results for PERCEPT-R datasets }
\end{table*}

\subsection{Error Bars}
\label{apendix:error-bars}
Tables \ref{tab:Global-Context1-WIDAR-eb} - \ref{tab:percept-eb} shows person-wise standard deviation values for generic $\MG$, conventionally finetuned $\MC$ and \textcolor{violet}{CRoP} $\MP$ models.

\begin{table*}[t]
\subfloat[Scenario 1 for WIDAR dataset]{
\resizebox{0.35\linewidth}{!}{
    \begin{tabular}{|c|cc|cc|cc|}
        \hline
        Model& \multicolumn{2}{|c|}{$\MG$} & \multicolumn{2}{|c|}{$\MP$ } &\multicolumn{2}{|c|}{$\MC$}\\
        \hline
         User& $\Ca$& $\Cu$& $\Ca$& $\Cu$& $\Ca$& $\Cu$ \\
         \hline
         0& 2.17 &0.88&3.69&0.49&2.09&0.29\\
         1& 1.49& 1.73& 1.94& 2.97&0.70&1.94\\
         2&3.61& 0.68&4.28&5.50&3.8&2.31\\
         \hline
         
    \end{tabular}}
    \label{tab:Global-Context1-WIDAR-eb}}
    \hfill
    \subfloat[Scenario 2 for WIDAR dataset]{
    \resizebox{0.35\linewidth}{!}{
    \begin{tabular}{|c|cc|cc|cc|}
        \hline
        Model& \multicolumn{2}{|c|}{$\MG$} & \multicolumn{2}{|c|}{$\MP$} &\multicolumn{2}{|c|}{$\MC$}\\
        \hline
         User& $\Ca$& $\Cu$& $\Ca$& $\Cu$& $\Ca$& $\Cu$ \\
         \hline
         0&2.67& 1.60&1.83& 0.95&1.02&0.41\\
         1& 1.58& 0.64&2.29& 0.97&2.22&0.68\\
         2& 2.49& 0.19&0.79&0.56&1.72&0.39\\
         \hline
    \end{tabular}}
    \label{tab:Global-Context2-WIDAR-eb}}

\vspace{6pt}
\subfloat[Scenario 1 for ExtraSensory dataset]{
\resizebox{0.35\linewidth}{!}{
    \begin{tabular}{|c|cc|cc|cc|}
        \hline
        Model& \multicolumn{2}{|c|}{$\MG$} & \multicolumn{2}{|c|}{$\MP$} &\multicolumn{2}{|c|}{$\MC$}\\
        \hline
         User& $\Ca$& $\Cu$& $\Ca$& $\Cu$& $\Ca$& $\Cu$ \\
         \hline
         61&2.63&0&4.14&1.65&3.28&2.52\\
         7C&1.45&0&1.17&0.81&1.36&0.83\\
         80&0.42&0&4.17&3.04&2.81&1.44\\
         9D&0.30&0&2.58&1.08&2.53&0.94\\
         B7&0.74&0&3.02&7.28&2.13&3.48\\
         \hline
    \end{tabular}}
    \label{tab:Global-Context1-extrasensory-eb}}
    \hfill
    \subfloat[Scenario 2 for ExtraSensory dataset]{
    \resizebox{0.35\linewidth}{!}{
    \begin{tabular}{|c|cc|cc|cc|}
        \hline
        Model& \multicolumn{2}{|c|}{$\MG$} & \multicolumn{2}{|c|}{$\MP$} &\multicolumn{2}{|c|}{$\MC$}\\
        \hline
         User& $\Ca$& $\Cu$& $\Ca$& $\Cu$& $\Ca$& $\Cu$ \\
         \hline
         61&3.93&0&4.03&2.67&4.72&1.00\\
         7C&2.45&0&2.35&2.50&2.85&0.45\\
         80&3.33&0&0.65&3.09&1.30&1.01\\
         9D&4.18&0&3.22&0.88&5.54&0.80\\
         B7&3.13&0&0.83&3.27&0.24&2.79\\
         \hline
    \end{tabular}}
    \label{tab:Global-Context2-extrasensory-eb}}

\vspace{6pt}
\subfloat[Scenario 1 for Stress Sensing - single context change]{
\resizebox{0.35\linewidth}{!}{
    \begin{tabular}{|c|cc|cc|cc|}
        \hline
        Model& \multicolumn{2}{|c|}{$\MG$} & \multicolumn{2}{|c|}{$\MP$} &\multicolumn{2}{|c|}{$\MC$}\\
        \hline
         User& $\Ca$& $\Cu$& $\Ca$& $\Cu$& $\Ca$& $\Cu$ \\
         \hline
         
         1&3.55&0&5.78 &4.53&7.94&6.17\\
         
         2& 5.13&0&14.65 & 3.71&8.74&10.49\\
         
         3&17.61&0&8.17 & 4.79&15.95&3.45\\
         \hline
         
    \end{tabular}}
    \label{tab:Global-Context1-EDA-1-eb}}
    \hfill
    \subfloat[Scenario 2 for Stress Sensing - single context change]{
    \resizebox{0.35\linewidth}{!}{
    \begin{tabular}{|c|cc|cc|cc|}
        \hline
        Model& \multicolumn{2}{|c|}{$\MG$} & \multicolumn{2}{|c|}{$\MP$} &\multicolumn{2}{|c|}{$\MC$}\\
        \hline
         User& $\Ca$& $\Cu$& $\Ca$& $\Cu$& $\Ca$& $\Cu$ \\
         \hline
         1& 4.95&0&3.02&0.66&3.02&2.34\\
         2&13.68&0&2.67&7.82&11.51&5.58\\
         3&8.25&0 &21.02&7.00&17.89&6.57\\
         \hline
    \end{tabular}}
    \label{tab:Global-Context2-EDA-1-eb}}

\vspace{6pt}
\subfloat[Scenario 1 for Stress Sensing - double context change]{
\resizebox{0.35\linewidth}{!}{
    \begin{tabular}{|c|cc|cc|cc|}
        \hline
        Model& \multicolumn{2}{|c|}{$\MG$} & \multicolumn{2}{|c|}{$\MP$} &\multicolumn{2}{|c|}{$\MC$}\\
        \hline
         User& $\Ca$& $\Cu$& $\Ca$& $\Cu$& $\Ca$& $\Cu$ \\
         \hline
         
         1&3.55 &0&5.75&3.33&7.94&4.78\\
         2& 5.13&0&14.66& 10.19&8.74&8.70\\
         3&17.61&0&8.17&5.52&15.95&3.58\\
         \hline
         
    \end{tabular}}
    \label{tab:Global-Context1-EDA-2-eb}}
    \hfill
    \subfloat[Scenario 2 for Stress Sensing - double context change]{
    \resizebox{0.35\linewidth}{!}{
    \begin{tabular}{|c|cc|cc|cc|}
        \hline
        Model& \multicolumn{2}{|c|}{$\MG$} & \multicolumn{2}{|c|}{$\MP$} &\multicolumn{2}{|c|}{$\MC$}\\
        \hline
         User& $\Ca$& $\Cu$& $\Ca$& $\Cu$& $\Ca$& $\Cu$ \\
         \hline
         1&4.95&0&3.02&3.00&3.02&2.52\\
         2&13.68&0&2.67&5.62&11.51&0.60\\
         3& 8.25&0&21.02&2.23&17.89&2.07\\
         \hline
         
    \end{tabular}}
    \label{tab:Global-Context2-EDA-2-eb}}

    \vspace{6pt}
\subfloat[PERCEPT-R]{
\resizebox{0.35\linewidth}{!}{
    \begin{tabular}{|c|cc|cc|cc|}
        \hline
        Model& \multicolumn{2}{|c|}{$\MG$} & \multicolumn{2}{|c|}{$\MP$} &\multicolumn{2}{|c|}{$\MC$}\\
        \hline
         User& $\Ca$& $\Cu$& $\Ca$& $\Cu$& $\Ca$& $\Cu$ \\
         \hline
         
         17&7.17&0&9.75&1.64&4.18&1.11\\
         25&1.34 &0&2.35&2.20&2.29&0.85\\
         28&3.38 &0&15.16&0.85&15.14&0.60\\
         336& 0&0&0&7.25&0&5.64\\
         344&5.50 &0&4.86&1.13&4.10&1.87\\
         361&15.75 &0&13.18&3.96&12.44&6.39\\
         362&3.10 &0&7.59&4.46&4.53&4.83\\
         55&3.75 &0&0&3.32&2.58&2.77\\
         586& 4.01&0&1.09&1.77&3.26&0.22\\
         587& 3.06&0&4.24&1.20&5.48&1.43\\
         589& 0.45&0&2.33&2.72&1.44&0.94\\
         590&2.26 &0&3.20&2.17&5.58&0.57\\
         591&3.36 &0&3.11&1.11&3.91&0.74\\
         61&9.30 &0&6.33&2.37&5.05&1.67\\
         67&5.50 &0&3.59&2.20&3.11&2.38\\
         80&6.11 &0&4.29&1.61&5.69&0.68\\
         \hline
         
    \end{tabular}}
    \label{tab:percept-eb}}
    \caption{Standard Deviation for Generic, conventionally finetunedd and CRoP models for WIDAR, ExtraSensory and  Stress Sensing dataset }

\end{table*}


\section{Impact of Pruning on different layer of the model}
\label{appendix:layer-wise-distance}
The overall impact of pruning can be attributed to two mechanims: the reguralization employed during the initial finetuning step (Algorithm \ref{alg:main} Line 2) and pruning employed during \emph{ToleratedPrune} step (Algorithm \ref{alg:main} Line 3). For this analysis, we use the LeNet model (shown in Figure \ref{fig:lenet-structure}) used for WIDAR dataset for all three users chosen for personalization. In order to capture the impacts of regularization, we compare average euclidean distance of two model states: initial finetuned and pruned with respect to the generic model  and per-layer pruning amount to discuss the impact of \emph{ToleratedPrune} module in Table \ref{tab:distance-pruning-generic}. Following \citet{li2017pruningfiltersefficientconvnets}, we avoid pruning of last decision making layer and show results for 3 convolution layers and first linear layers.  Based on the distances and pruning amounts shown in Table \ref{tab:distance-pruning-generic} , we can conclude that for all users the first linear layer and the third convolution layer  undergo the highest amount of change during these steps.  We explain this pattern as follows.


It is important to note that the initial convolution layers capture the generic traits in the form of low level features while deeper layer capture high level features which are task-specific \citep{RIBEIRO201813}. In the LeNet model, the third convolution layer extracts high level features, while the first linear layer identifies the importance of these features and represents this information in the form of its weight magnitudes \citep{Goodfellow2016}. Thus, during the initial finetuning stage, these two layers undergo the highest amount of change in order to learn traits specific to a person's available context data. This is inline with the distances shown in Table \ref{tab:distance-pruning-generic} as the distance between these layers of the initial finetuned and the generic model are the highest. Additionally, the regularization posed by the initial finetuning stage constrains the learned parameters to a subset of these layers while penalizing the parameters which were less important of the user-specific traits present in available context $\Ca$ data. This allows \emph{ToleratedPrune} module to impose higher pruning at these layers.

Thus, it can be concluded that \textcolor{violet}{CRoP} implicitly retains generic traits in initial feature extraction layer by minimally impacting them while allowing the deeper layers to learn traits specific to target users' available context data. Simultaneously, it constrains the learning of these new traits to smaller subset of the deeper layers, in order to keep vacancy to bring back the generic trends presents in the generic model.

\begin{table}
 
    \centering
    \resizebox{0.99\linewidth}{!}{
    \begin{tabular}{|c|c|c|c||c|c|c||c|c|c|}
    \hline
         Euclidean Distance& \multicolumn{3}{c||}{User0} & \multicolumn{3}{c||}{User1} &\multicolumn{3}{c|}{User2} \\
         \hline
          Generic model&Initial Finetuned& Pruned  & Pruning \% &Initial Finetuned & Pruned   & Pruning \% &Initial Finetuned& Pruned  & Pruning \% \\
         \hline
         First Convolution Layer & 1.63 & 2.87  & 25.9 &  2.69&5.47  &  26.2  &   6.04  & 18.35  & 25.8\\
         Second Convolution Layer&0.99 & 1.90  & 15.1&    1.75 & 3.96 & 20.9  &   10.29& 23.94 & 46.7\\
         Third Convolution Layer &1.67 &3.70  &  22.4
&    1.89 & 6.82 &30.3  &  17.98 & 50.70&64.6\\
         First Linear Layer & 45.42 & 55.89 & 49.2 & 
  44.40& 70.80 & 60.1  &    65.00  & 212.04 & 96.0\\
         \hline
    \end{tabular}}

\caption{Average euclidean distance ($X 10^ {-6}$) per parameter for different model states as compared to the generic model, and pruning amount of different layers of the LeNet Model  for three users chosen for personalization for WIDAR dataset}
\label{tab:distance-pruning-generic}
\end{table}





\end{document}